%% file: main.tex
\documentclass[10pt,twocolumn,letterpaper]{article}

\usepackage[pagenumbers]{iccv} 

\input{preamble}
\usepackage{overpic}
\usepackage{wrapfig}
\usepackage{amsmath}
\usepackage{empheq}
\usepackage{comment}
\usepackage{makecell}
\usepackage{multirow}
\usepackage{multicol}
\usepackage{float}
\usepackage{marvosym}

\definecolor{iccvblue}{rgb}{0.21,0.49,0.74}

\usepackage[pagebackref=iccvblue,breaklinks,colorlinks,citecolor=iccvblue,linkcolor=iccvblue]{hyperref}

\def\paperID{7047} 
\def\confName{ICCV}
\def\confYear{2025}

\title{ResGS: Residual Densification of 3D Gaussian for Efficient Detail Recovery}

\author{Yanzhe Lyu \quad Kai Cheng  \quad Xin Kang \quad Xuejin Chen$^\dagger$ \\
MoE Key Laboratory of Brain-inspired Intelligent Perception and Cognition\\
University of Science and Technology of China\\}

\begin{document}
\twocolumn[{
    \renewcommand\twocolumn[1][]{#1}
    \maketitle
    \vspace{-2.0mm}
    \begin{center}
    \vspace{-3.0mm}
        \captionsetup{type=figure}
        \includegraphics[width=\textwidth]{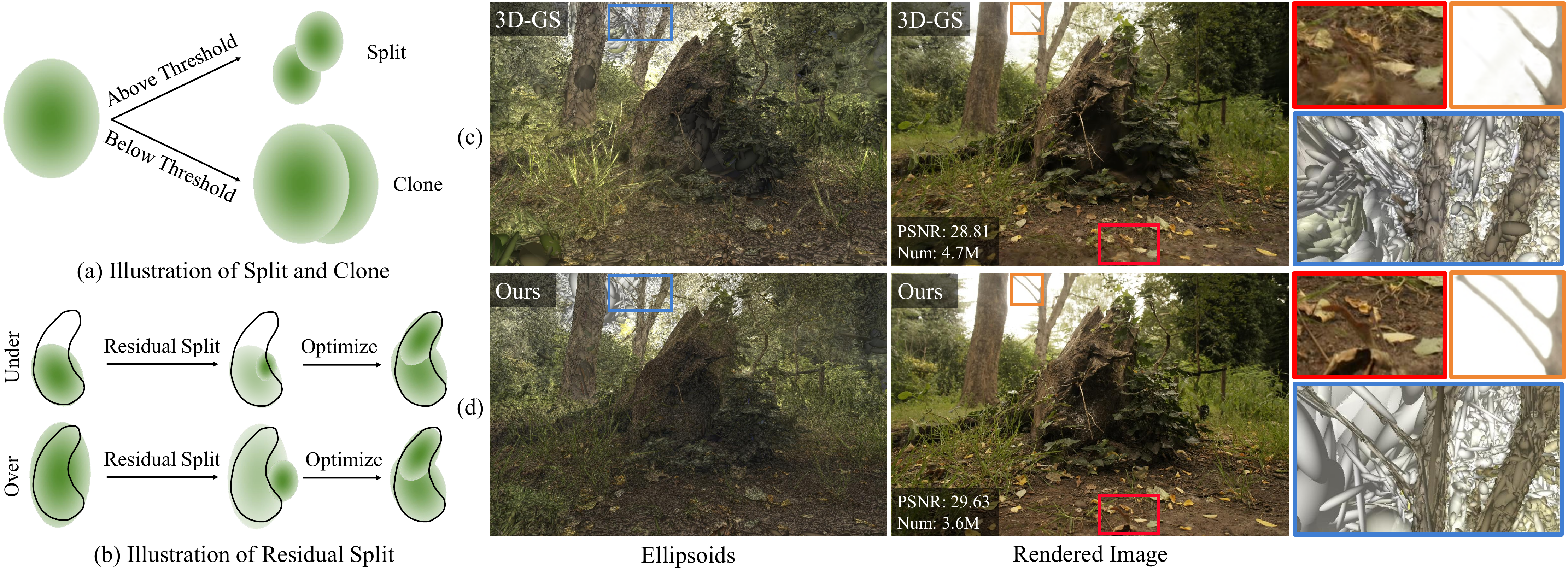}
        \captionof{figure}{
        We propose \modelname, a pipeline to boost the rendering quality of 3D-GS \cite{Kerbl20233DGS} while improving efficiency. 
        (a) The 3D-GS densification operations, split and clone, rely on binary selection and cannot adaptively address the challenges they encounter.
        (b) Our \modelname~employs a single densification operation, \densifyname, that adaptively addresses under-reconstruction and over-reconstruction. 
        (c) 3D-GS tends to generate Gaussians with sizes in small variation and low rendering quality.
        (d) From the same initialization, our method can \textcolor{red}{capture fine details} and \textcolor{orange}{retrieve sufficient geometry} effectively while \textcolor{blue}{reducing redundancy}. 
        }
        \label{fig:teaser}
    \end{center}
    }]

\footnotetext{
\llap{\textsuperscript{$\dagger$}}Corresponding author.
}

\maketitle
\input{sec/0_abstract}    
\input{sec/1_intro}
\input{sec/2_related_work}

\input{sec/3_method}

\input{sec/4_experiments}
\input{sec/5_conclusion}
\input{sec/6_ack}
{
    \small
    \bibliographystyle{ieeenat_fullname}
    \bibliography{main}
}
\input{sec/X_suppl}


\end{document}

%% file: preamble.tex
\usepackage[dvipsnames]{xcolor}
\usepackage{colortbl}

\definecolor{placeholder}{rgb}{0.6,0.8,0.95}

\newcommand{\modelname}{ResGS}
\newcommand{\densifyname}{residual split}

\definecolor{amber}{rgb}{1.0, 0.75, 0.0}

\definecolor{visible-blue}{rgb}{0.286, 0.525, 0.910}

\definecolor{tabfirst}{rgb}{1, 0.7, 0.7} 
\definecolor{tabsecond}{rgb}{1, 0.85, 0.7} 
\definecolor{tabthird}{rgb}{1, 1, 0.7} 
\definecolor{blue}{rgb}{0.251, 0.498, 0.824}

%% file: sec/0_abstract.tex
\begin{abstract}

Recently, 3D Gaussian Splatting (3D-GS) has prevailed in novel view synthesis, achieving high fidelity and efficiency. However, it often struggles to capture rich details and complete geometry. 
Our analysis reveals that the 3D-GS densification operation lacks adaptiveness and faces a dilemma between geometry coverage and detail recovery.
To address this, we introduce a novel densification operation, \textbf{\densifyname}, which adds a downscaled Gaussian as a residual. Our approach is capable of adaptively retrieving details and complementing missing geometry. To further support this method, we propose a pipeline named \modelname. Specifically, we integrate a Gaussian image pyramid for progressive supervision and implement a selection scheme that prioritizes the densification of coarse Gaussians over time. Extensive experiments demonstrate that our method achieves SOTA rendering quality. 
Consistent performance improvements can be achieved by applying our \densifyname~on various 3D-GS variants, underscoring its versatility and potential for broader application in 3D-GS-based applications. 
Project page: \href{https://yanzhelyu.github.io/resgs.github.io/}{https://yanzhelyu.github.io/resgs.github.io/}.

\end{abstract}

%% file: sec/1_intro.tex
\section{Introduction}
\label{sec:intro}

\begin{figure*}[t!]
\includegraphics[width=\linewidth]{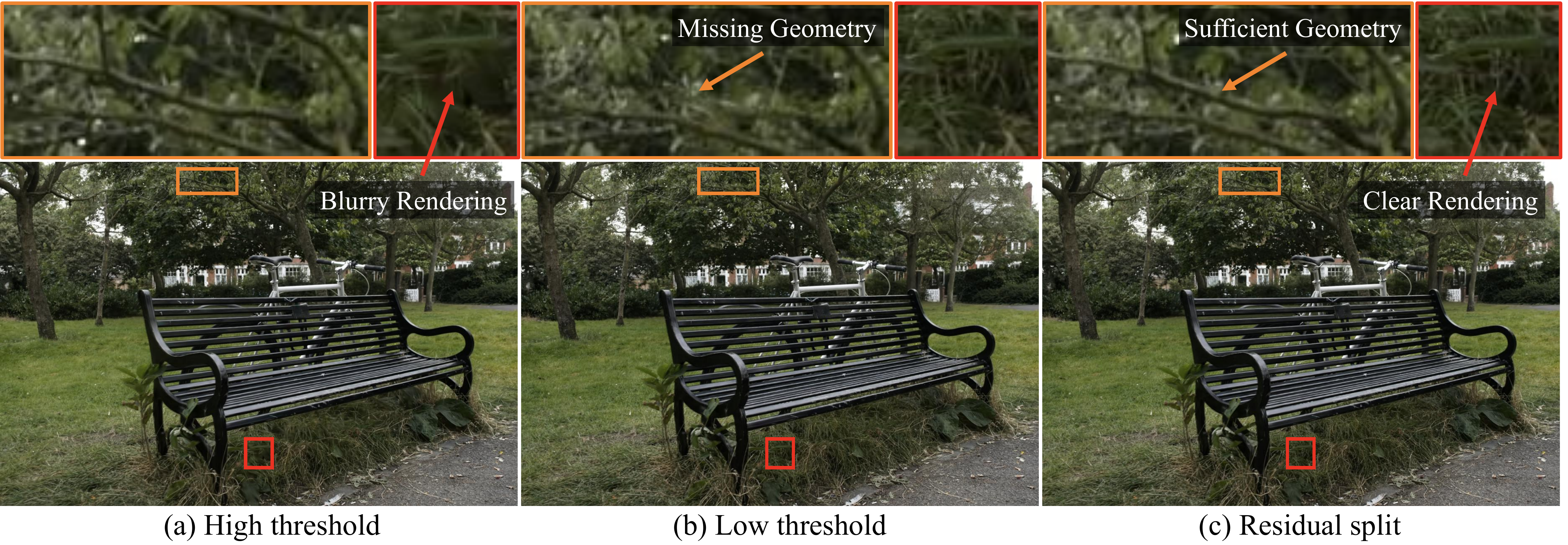}
\caption{
\textbf{An illustration of the limitation of split and clone.} 
\textbf{(a) selects between split and clone based on a high threshold, (b) a low threshold, and in (c) the densification operation is replaced as \densifyname.} 
We provide visualizations of the final rendered image. A high threshold ensures sufficient structural coverage but causes over-reconstruction, leading to blurry rendering. Conversely, a low threshold mitigates blurry rendering but results in under-reconstruction, compromising coverage and creating a difficult trade-off. Our method, however, adaptively tackles both issues without facing this trade-off.}

\label{fig:3DGS_limitation}
\end{figure*}

\label{sec:3dgs_limitation}

Novel View Synthesis (NVS) has continued to be a hot topic throughout the years. Traditional methods like point clouds and meshes \cite{BotschEurographicsSO, Goesele2007MultiViewSF, Chen1993ViewIF} can achieve high rendering speed but often lack fidelity. On the other hand, neural volumetric approaches \cite{Mildenhall2020NeRF, Barron2021MipNeRFAM, Mller2022InstantNG} can achieve high fidelity for novel views 
but significantly sacrifice rendering efficiency. 
Recently, 3D-GS \cite{Kerbl20233DGS} has shown both high fidelity and fast rendering speed for novel view synthesis.  
However, 3D-GS still suffers from setbacks, often failing to capture fine details or recover complete geometry, as shown by the red and orange rectangles in Fig.~\ref{fig:teaser} and Fig.~\ref{fig:3DGS_limitation}. 

While various approaches \cite{Zhang2024FreGS3G, zhang2024pixelgs, Ye2024AbsGSRF, Fang2024MiniSplattingRS} have been proposed to tackle these challenges and have significantly advanced the rendering quality of 3D-GS-based methods to state-of-the-art (SOTA) levels, they seldom adapt the densification operation of 3D-GS. In this paper, we go through the densification procedure and propose a novel adaptive densification approach to improve reconstruction quality.

The densification operation used in 3D-GS-based methods consists of two operations: split and clone. The split operation is designed to address over-reconstruction, where a single Gaussian covers too much area, and the clone operation aims to compensate for under-reconstruction. However, selecting between the two operations properly in different areas is challenging and existing methods rely on a threshold-based decision mechanism. As illustrated in Fig.~\ref{fig:teaser} (a), Gaussians whose scale $s=max(s_x, s_y, s_z)$ is larger than a predefined threshold $\tau_s$ undergo split, while those below it use clone. The simplicity of this threshold-based densification usually leads to suboptimal results and encounters a dilemma in threshold selection. As highlighted by the red rectangles in Fig.~\ref{fig:3DGS_limitation} (a), a high threshold results in insufficient split preventing the model from capturing fine details and leading to blurry rendering. On the other hand, a low threshold suppresses clone operations, leading to missing geometry, as illustrated in the orange boxes in Fig.~\ref{fig:3DGS_limitation} (b). Furthermore, 3D-GS tends to first split and then clone, resulting in objects being fitted with Gaussians of similar scales. As highlighted by the blue rectangles in Fig.~\ref{fig:teaser}, 3D-GS fits textureless areas with an excessive number of small Gaussians, which is redundant.

To tackle the aforementioned problems, we propose \textbf{\densifyname}, a novel densification operation in 3D Gaussian splatting. As illustrated in Fig.~\ref{fig:teaser} (b), the core idea is to spawn a downscaled replica as a residual component while reducing the original Gaussian’s opacity for any under-optimized Gaussian. This simultaneously expands spatial coverage and injects finer details, eliminating the dilemma of selecting split or clone.

However, directly applying residual split does not produce significant improvements. This is due to the inefficiency the 3D-GS initialization. At early iterations, the scene is represented by a few coarse Gaussians that very likely face both over-reconstruction and under-reconstruction. Gaussians added through densification are challenging to optimize, as they must address both issues simultaneously. It's desired to develop an approach that decouples coverage and detail optimization.

Correspondingly, we propose a coarse-to-fine training pipeline, \textbf{ResGS}, that progressively shifts the focus of training, facilitating optimization. 
Firstly, we structure the training process into stages, employing multi-scale images for gradual supervision, ensuring that the early stages focus on the overall structure and the later stages focus on details. Secondly, we develop a progressive Gaussian selection scheme for densification, which gradually encourages coarser Gaussians to undergo further refinement. Therefore, finer structures are introduced at later iterations, avoiding premature optimization. These two techniques simplify optimization and thereby improve rendering quality.

Our contributions are summarized as follows: \begin{itemize}
\item  We propose a novel densification operation for 3D Gaussians based on a residual strategy to overcome the limitations of fixed threshold for determining split or clone in 3D-GS; 

\item  We develop a coarse-to-fine training pipeline, ResGS. Equipped with the proposed residual split, it incorporates a multi-scale supervision with image pyramids and a Gaussian selection scheme to enhance rendering quality;

\item 
Our method achieves SOTA rendering quality in NVS on three datasets, including Mip-NeRF360 \cite{Barron2021MipNeRFAM}, Tanks\&Temples \cite{Knapitsch2017TanksAT}, and DeepBlending \cite{Hedman2018DeepBF}. Furthermore, applying our proposed \densifyname~to multiple 3D-GS variants consistently brings performance improvements, demonstrating that our new densification operation is more effective than densification with a fixed scale threshold. 

\end{itemize}

%% file: sec/2_related_work.tex
\section{Related Work}
\label{sec:related_work}

\subsection{Novel View Synthesis}

Novel view synthesis (NVS) aims to acquire novel view images for a scene through a set of multi-view images. %
As a pioneer, NeRF \cite{Mildenhall2020NeRF} introduced an MLP-based implicit representation. 
More specifically, it uses an MLP to predict the color and opacity of a point in a scene and uses differentiable volume rendering to render images. Due to its photo-realistic quality, it has been extensively studied \cite{Chen2021MVSNeRFFG, Jain2021PuttingNO, Chan2022, Zhang2022FDNeRFFD, Kwak2023GeCoNeRFFN, Chen2023PAniC3DSS, Barron2021MipNeRFAM, Barron2023ZipNeRFAG}. However, NeRF-based methods often suffer from long training and rendering time. Therefore, many works \cite{Reiser2021KiloNeRFSU, Reiser2023MERFMR, Hedman2021BakingNR, Hu_2022_CVPR} have been proposed to extend NeRF into real-time rendering.

Recently, 3D-GS \cite{Kerbl20233DGS} has stirred an evolution in the field by achieving both high-fidelity and real-time rendering. Due to its high performance, many works have enabled 3D-GS into many downstream tasks such as SLAM \cite{Matsuki2023GaussianSS, Yan2023GSSLAMDV}, 3D generation 
 \cite{Chen2023Textto3DUG, Tang2023DreamGaussianGG, Yi2023GaussianDreamerFG, Zou2023TriplaneMG}, 4D scene modeling \cite{Yang2023RealtimePD, Wu20234DGS, Guo2024Motionaware3G, Kratimenos2023DynMFNM}, and surface reconstruction \cite{Huang2DGS2024, Guedon2023SuGaRSG}.

However, the original 3D-GS still has limitations, and myriad works have been proposed to increase the render quality and efficiency of 3D-GS. Mip-Splatting \cite{Yu2023MipSplattingA3} introduced a 2D dilation filter and a 3D smoothing filter to mitigate aliasing and artifacts caused by changes in focal length. Scaffold-GS \cite{Lu2023ScaffoldGSS3} proposed the concept of neural Gaussians which is to tie Gaussians around an anchor and use an MLP to predict its attributes. Thus, the model is more robust to significant view changes and can improve rendering quality.
AbsGS \cite{Ye2024AbsGSRF} discovered that when accumulating the view space gradients used in densification through pixels, there could be a collision, for gradients at various pixels might have different positivities. 
They applied an absolute function before summing through pixels to mitigate the issue.  
Pixel-GS \cite{zhang2024pixelgs} analyzed 3D-GS artifacts from the pixel perspective and introduced a method to select Gaussians for densification based on this analysis. GaussianPro \cite{Cheng2024GaussianPro3G} introduced a progressive propagation strategy to propagate Gaussians to under-initialized areas. To reduce redundancy and increase rendering quality, Mini-Splatting \cite{Fang2024MiniSplattingRS} proposed a densification and simplification algorithm to reorganize Gaussians' spatial positions. 

With the emergence of numerous works focused on optimizing NeRF and 3D-GS, progressive training schedules—proven effective in multiple scenarios \cite{Bengio2009CurriculumL, He2014SpatialPP, Rusu2016ProgressiveNN, Karras2017ProgressiveGO} have earned attention. 

\subsection{Progressive Training in NVS}
Several progressive training schedules have been explored upon NeRF-based methods. BungeeNeRF \cite{Xiangli2021BungeeNeRFPN} captured different levels of details in large city-scale datasets by gradually adding closer views to the training views and appending residual blocks to the MLP throughout the training process. Pyramid NeRF \cite{Zhu2023PyramidNF} employs a coarse-to-fine strategy, using an image pyramid to progressively introduce high-frequency details while simultaneously subdividing the scene representation.

Recently, with the prevalence of 3D-GS, many progressive training pipelines have been proposed. Octree-GS \cite{Ren2024OctreeGSTC} proposed an Octree-based 3D-GS approach, and the Octree depth is progressively deepened. FreGS \cite{Zhang2024FreGS3G} designed a progressive frequency regulation for 3D-GS which is to calculate the loss of rendered images in the frequency domain and gradually add higher frequency components.

%% file: sec/3_method.tex
\section{Our Method}
\begin{figure*}[t!]
\includegraphics[width=\linewidth]{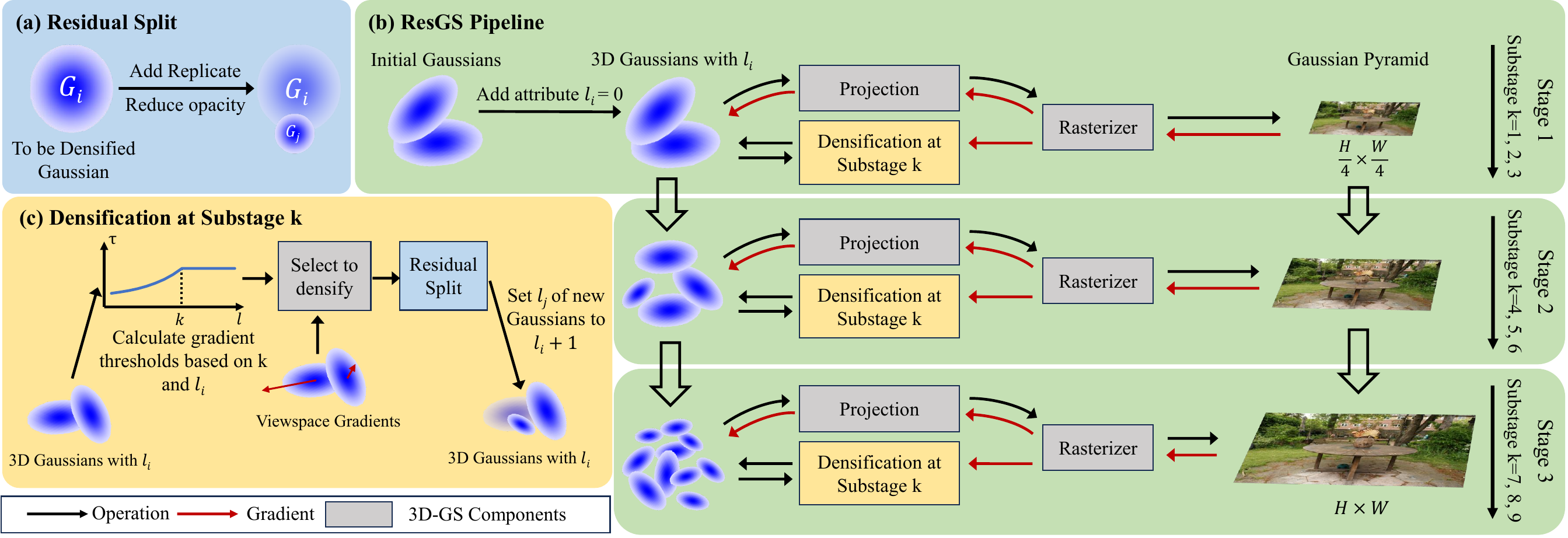}
\caption{
\textbf{Overview of our \modelname\ pipeline.} (a) The core of our pipeline, \densifyname, involves adding a downscaled replicate and then reducing the opacity of the original Gaussian. (b) We assign initial Gaussians a temporary attribute $l_i=0$ for densification selection, which is discarded after training. Next, the pipeline is split into $L$ stages,  with each stage trained on images downscaled using an image pyramid. Each single stage is further divided evenly into $K$ substages, for selecting Gaussians to densify. (c) The points selected for densification are determined by the substage $k$, $l_i$, and viewspace gradients of Gaussians. 
Residual split is performed upon selected Gaussians, and $l_j$ would be assigned to new Gaussians.
}
\label{fig:overview}
\end{figure*}

\subsection{Preliminaries}
\label{sec:prelim}
\paragraph{3D Gaussian Splatting.}
The 3D-GS work \cite{Kerbl20233DGS} represents a 3D scene with a set of anisotropic 3D Gaussians for fast and high-quality rendering.
Each Gaussian primitive consists of two parameters, mean $\boldsymbol{\mu}$ and covariance matrix $\mathbf{\Sigma}$, representing its position and shape:
\begin{equation}
    \begin{aligned}
        &G(\mathbf{x}) = e^{-\frac{1}{2}(\mathbf{x}-\boldsymbol{\mu})^{T} \mathbf{\Sigma}^{-1} (\mathbf{x}-\boldsymbol{\mu})}.
    \end{aligned}
\end{equation}
To model the positive semi-definite property of the covariance matrix $\mathbf{\Sigma}$, a scaling matrix $\mathbf{S}$ and a rotation matrix $\mathbf{R}$ is used as:
\begin{equation}
    \begin{aligned}
        &\mathbf{\Sigma} = \mathbf{RS}\mathbf{S}^T\mathbf{R}^T,
    \end{aligned}
\end{equation}
where a 3D vector $\mathbf{s}=({s}_x,{s}_y,{s}_y)$ is used for the scaling matrix $\mathbf{S}$. 

Furthermore, each Gaussian also contains a spherical harmonics coefficients $SH$ and opacity $o$. 
In rendering, tile-based rasterization is applied: 3D Gaussians $G(\mathbf{x})$
are projected as 2D Gaussians $G'(\mathbf{x})$ on the image plane. A view-dependent color $c$ is then derived from $SH$, and the 2D Gaussians are sorted by depth and rendered using alpha-blending:
\begin{equation}
    \begin{aligned}
        &C(\mathbf{x}) = \sum_{i \in N_\mathbf{x}} c_i \alpha_i(\mathbf{x}) \prod_{j=1}^{i-1} (1-\alpha_j(\mathbf{x})),\\
        &\alpha_i(\mathbf{x}) = c_i G_i'(\mathbf{x}),
    \end{aligned}
\end{equation}
where $\mathbf{x}$ is a pixel position in the rendered image, $C(\mathbf{x})$ the color of the pixel, and $N_\mathbf{x}$ the number of Gaussians that covers $\mathbf{x}$. 

During the training process, 3D Gaussians are periodically densified. 3D-GS calculates the average viewspace gradients $\Delta L_i$ over a certain number (e.g. 100) of iterations to determine whether to perform densification. If $\Delta L_i$ exceeds threshold $\tau$, a split or clone action is triggered.
The split operation addresses over-reconstruction, where excessive space is covered, while clone addresses under-reconstruction, where additional spatial coverage is needed. 
To determine whether to split or clone, 3D-GS \cite{Kerbl20233DGS} sets a threshold $\tau_s=kR$, where $k$ is a predefined factor and $R$ is the scene extent.
Split occurs when the scale $s=max({s}_x,{s}_y,{s}_y)$ of a Gaussian surpasses the threshold $\tau_s$. Clone is applied when $s<\tau_s$. 


\subsection{\modelname}

We propose \modelname, a coarse-to-fine pipeline to boost the rendering quality of 3D-GS with a novel residual densification operation. Fig.~\ref{fig:overview} shows the overview of our method. 
We propose a novel densification operation, \densifyname, (Sec.~\ref{sec:densification}), to tackle the limitations of the 3D-GS densification operation. Furthermore, we introduce a split-stage supervision schedule to add information to the scene gradually~(Sec.~\ref{sec:GSP}). 
We also design a Gaussian selection scheme that progressively encourages densification for coarse Gaussians to enhance details, as described in Sec.~\ref{sec:vary_threshold}.

\subsubsection{Residual Split}
\label{sec:densification}
The core idea of our densification operation is to generate downscaled Gaussians as residuals to supplement those requiring densification. The newly added Gaussian adaptively enhances the reconstructed model, allowing the scene to be refined as needed. The method is named as \densifyname. As illustrated in Fig.~\ref{fig:overview} (a), for a to-be-densified Gaussian $G_i$, our densification operation consists of two steps. Firstly, we add a down-scaled replicate $G_j$. 
Specifically, its opacity $o_j$, rotation $\mathbf{R}_j$, spherical harmonics coefficients $SH_j$ would be same as $G_i$, $\boldsymbol{\mu}_j$ is sampled using the original 3D Gaussian as a PDF for sampling, and $\mathbf{S}_j$ is defined as $\mathbf{S}_i$ divided by a predefined factor $\lambda_s$:
\begin{equation}
    \begin{aligned}
        &\mathbf{S}_j = \frac{1}{\lambda_s}\mathbf{S}_i,\\
        &\mathbf{R}_j = \mathbf{R}_i,\ SH_j = SH_i,\ o_j = o_i,\\
        &\boldsymbol{\mu}_j \sim \mathcal{N}(\boldsymbol{\mu}_i, \boldsymbol{\Sigma}_i),
    \end{aligned}
\end{equation}
where $\boldsymbol{\Sigma}_i$ is the covariance matrix of $G_i$. 
Then, to tackle the density increase in the overlap of the two Gaussians, we reduce the opacity of the original Gaussian. This is accomplished by multiplying its opacity by a predefined factor $\beta$:
%
\begin{equation}
    \begin{aligned}
        &o'_i = \beta o_i.
    \end{aligned}
\end{equation}

\begin{figure}[t!]
\includegraphics[width=\linewidth]{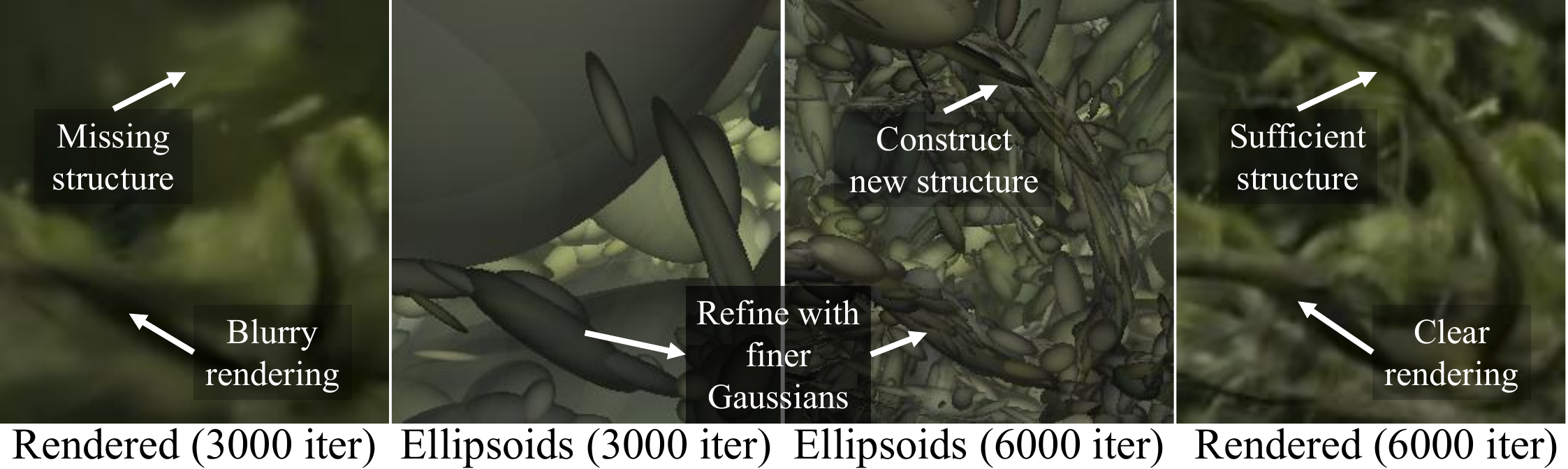}
\caption{
\textbf{Intermediate results after applying \densifyname.} We saved the Gaussians at 3,000 and 6,000 iterations after applying residual split. As observed, residual split can adaptively handle over-reconstruction and under-reconstruction by refining blurry regions with finer Gaussians while also effectively reconstructing missing structures. 
} 
\label{fig:intermediate_results}
\end{figure}

While the clone and split in the original 3D-GS pipeline lack adaptiveness, our \densifyname~can fill in missing geometry and recover details adaptively without requiring a trade-off, as shown in Fig.~\ref{fig:3DGS_limitation} (c). We further illustrate this using intermediate results, as shown in Fig.~\ref{fig:intermediate_results}.
In areas where details are challenging to recover, down-scaled Gaussians of finer granularity are progressively introduced to refine and reveal intricate details. 
On the other hand, since densification does not change the scale of the original Gaussian, the total coverage is increased after densification, helping to reconstruct missing structures. 
Moreover, our method’s approach to generate nonidentical Gaussians of varying scales enables efficient shape fitting. 
As illustrated by the blue rectangles in Fig.~\ref{fig:teaser}, this non-identicalness allows us to fill textureless regions with a few large Gaussians, significantly reducing redundancy.
Consequently, our approach delivers higher rendering quality while using fewer Gaussians.

\subsubsection{Image Pyramid as Supervision}
\label{sec:GSP}


To decouple coverage and detail optimization, we design a coarse-to-fine supervision schedule to gradually integrate fine high-frequency information into the scene. 
%
Specifically, based on the original multi-view training pictures, we build an image pyramid $\{\mathcal{I}_i\}_{i=1}^{L}$ consisting of $L$ layers. 
$\mathcal{I}_i$ corresponds to the set of images at different viewpoints in layer $i$. $\mathcal{I}_L$ is the set of original pictures:
\begin{equation}
    \begin{aligned}
        &\mathcal{I}_i = \{ P^v_i \}_{v \in V},
    \end{aligned}
\end{equation}
where $V$ is the set of viewpoints and $P^v_i$ refers to the image at viewpoint $v$ in the $i^{\text{th}}$ layer. 
Let $(H_L^v,W_L^v)$ be the resolution of $P_L^v$, and $(H_i^v,W_i^v)$ represent the resolution of $P_i^v$. 
Then there is:
\begin{equation}
    \begin{aligned}
        &(H_i^v,W_i^v)=(H_L^v/2^{L-i},W_L^v/2^{L-i}).
    \end{aligned}
\end{equation}

Correspondingly, our training process is divided into $L$ stages. At the $i$-th stage, the Gaussians are trained under the supervision of $\mathcal{I}_i$.
At the initial stage with lower image resolution, the 3D Gaussians are optimized to construct an overall structure of the whole scene from coarse and low-frequency features.
As the image resolution increases, high-frequency details of the scene gradually emerge, shifting the focus toward adding finer, more intricate features.

\subsubsection{Varying Gradient Threshold}
\label{sec:vary_threshold}

To further emphasize the progressive training, we modify the strategy for selecting Gaussians to densify. 
As the training proceeds, the training focus is desired to shift toward finer details. Thus, we encourage the densification of coarse Gaussians to capture more detail over time. 
In our \densifyname, the generated Gaussians are finer than the original ones. 
Thus, we can track the relative fineness level of each Gaussian $G_i$, denoted as $l_i$. 
We assign $l_i$ of all initial Gaussians to 0. 
If the level of the Gaussian being densified is $l_i$, we assign the newly added Gaussian a level of $l_i + 1$. Thus, as $l_i$ gets higher, Gaussians tend to be smaller and finer. 
Note that $l_i$ is only utilized during rendering and discarded after training.

A single stage is then divided evenly into $K$ substages, creating $L\times K$ substages in total. 
The gradually increasing substage can be seen as an indicator of the scene details level. 
At substage $k$, we promote additional densification for Gaussians with $l_i < k$ and prioritize those at lower levels, achieved by applying varying gradient thresholds. Specifically, the threshold $\tau_{k,i}$ for deciding whether densification is needed for a Gaussian $G_i$ is set as:

\begin{equation}
\tau_{k,i} = \left\{
\begin{aligned}
    & \tau, l_i \geq k, \\
    & \frac{\tau}{\alpha^{k-l_i}}, l_i < k,
\end{aligned}
\right.
\end{equation}
where $\tau$ and $\alpha > 1$ are predefined values.

%% file: sec/4_experiments.tex
\section{Experiments and Results}

\begin{table*}[]
\centering
\caption{\textbf{Quantitative comparison to previous methods on real-world datasets.} Our method achieves SOTA in the Mip-NeRF360 dataset and comparable results with SOTA on the Tanks\&Temples and Deep Blending datasets.
$^*$ indicates results generated by retrained models for different rendering resolution.
}
\label{tab:quality}
\resizebox{\linewidth}{!}{
\begin{tabular}{c|cccc|cccc|cccc}
\toprule
Dataset & \multicolumn{4}{c|}{Mip-NeRF360} & \multicolumn{4}{c|}{Tanks\&Temples} & \multicolumn{4}{c}{Deep Blending} \\
\begin{tabular}{c|c} Method & Metrics \end{tabular}  & PSNR \(\uparrow\) & SSIM \(\uparrow\) & LPIPS \(\downarrow\)& Mem. & PSNR \(\uparrow\) & SSIM \(\uparrow\) & LPIPS \(\downarrow\) & Mem.  & PSNR \(\uparrow\) & SSIM \(\uparrow\) & LPIPS \(\downarrow\) & Mem.  \\
\midrule
\textbf{Plenoxels}~ & 23.08 & 0.626 & 0.463 &  2.1GB   & 21.08 & 0.719 & 0.379 & 2.3GB & 23.06 & 0.795 & 0.510 & 2.7GB \\ 
\textbf{Instant-NGP}~ & 25.59 & 0.699 & 0.331 &  48MB   & 21.92 & 0.745 & 0.305 & 48MB & 24.96 & 0.817 & 0.390 & 48MB \\ 
\textbf{Mip-NeRF360}~ & 27.69 & 0.792 & 0.237 &  8.6MB   & 22.22 & 0.759 & 0.257 & 8.6MB & 29.40 & 0.901 & 0.245 & 8.6MB \\ 
\hline

\textbf{Scaffold-GS$^*$} ~ & 27.69 & 0.812 & 0.225 & 176MB  & 23.96 & 0.853 & 0.177 & 87MB  & \cellcolor{tabsecond}30.21 & \cellcolor{tabsecond}0.906 & 0.254 & 66MB  \\ 

\textbf{Octree-GS$^*$} ~ & 27.52 & 0.812 & 0.218 & 124MB & \cellcolor{tabfirst}24.52 & \cellcolor{tabsecond}0.866 & 0.153 & 84MB  & \cellcolor{tabfirst}30.41 & \cellcolor{tabfirst}0.913 & 0.238 & 93MB  \\ 
\hline

\textbf{3D-GS} ~ & 27.21 & 0.815 & 0.214 & 734MB  & 23.14 & 0.841 & 0.183 & 411MB & 29.41 & 0.903 & 0.243 &676MB \\ 

\textbf{AbsGS} ~ & 27.49 & 0.820 & 0.191 & 728MB  & 23.73 & 0.853 & 0.162 & 304MB  & 29.67 & 0.902 & 0.236 & 444MB  \\ 
\textbf{Pixel-GS$^*$} ~ & 27.52  & 0.822 & 0.191 & 1.32GB  & 23.78 & 0.853 & 0.151 & 1.06GB  & 28.91 & 0.892 & 0.250 & 1.09GB  \\

\textbf{FreGS} ~ &  \cellcolor{tabthird}27.85 & 0.826 & 0.209 & -  & 23.96 & 0.849 & 0.178 & -  &  29.93 &  \cellcolor{tabthird}0.904 & 0.240 & -  \\

\textbf{Mini-Splatting-D} ~ &  27.51 & \cellcolor{tabsecond}0.831 & \cellcolor{tabsecond}0.176 & 1.11GB  & 23.23 & 0.853 & \cellcolor{tabsecond}0.140 & 1.01GB  &  29.88 &  \cellcolor{tabsecond}0.906 & \cellcolor{tabfirst}0.211 & 1.09GB  \\

\hline
\textbf{Ours-Small (AbsGS)} &  \cellcolor{tabsecond}27.94 & \cellcolor{tabthird}0.830 &0.191 & 342MB  & \cellcolor{tabthird}24.33 &0.862 & 0.150 & 187MB  & \cellcolor{tabthird}30.01 & \cellcolor{tabsecond}0.906 & 0.234 & 289MB  \\

\textbf{Ours (3D-GS)} &  \cellcolor{tabfirst}28.00 & \cellcolor{tabsecond}0.831 &\cellcolor{tabthird}0.187 & 600MB  & 24.26 & \cellcolor{tabthird}0.865 & \cellcolor{tabthird}0.141 & 354MB  & 29.93 & 0.903 & \cellcolor{tabthird}0.232 & 596MB  \\

\textbf{Ours (AbsGS)} & \cellcolor{tabfirst}28.00 & \cellcolor{tabfirst}0.833 & \cellcolor{tabfirst}0.174 & 698MB  & \cellcolor{tabsecond}24.38 & \cellcolor{tabfirst}0.867 &\cellcolor{tabfirst}0.132 & 351MB  & 29.91 & 0.902 & \cellcolor{tabsecond}0.227 & 586MB  \\

\bottomrule
\end{tabular}}
\end{table*}

\subsection{Experimental Settings}
\label{sec:exp_setting}
Following previous works \cite{Kerbl20233DGS, Ye2024AbsGSRF, Zhang2024FreGS3G,Lu2023ScaffoldGSS3, Ren2024OctreeGSTC, Fang2024MiniSplattingRS}, evaluation of our model was done on three datasets, including all nine scenes from Mip-NeRF360 \cite{Barron2021MipNeRFAM}, two scenes from Tanks\&Temples \cite{Knapitsch2017TanksAT}, and two scenes from Deep Blending \cite{Hedman2018DeepBF}. The test and train splits follow those of the original 3D-GS \cite{Kerbl20233DGS}, and the image resolution is also maintained as in 3D-GS.
Three commonly used metrics were used to quantify the fidelity of rendered images: PSNR, SSIM \cite{Wang2004ImageQA}, and LPIPS \cite{Zhang2018TheUE}. We also report the memory consumption as an indicator of the number of Gaussians. 

We trained three versions of our model: two with AbsGS \cite{Ye2024AbsGSRF}, leveraging its enhanced indication of Gaussians for densification—one standard version and one small version with higher gradient thresholds—and a third version using 3D-GS as the baseline. We applied the periodic opacity reduction from AbsGS across all three versions, as it proved to be a more effective alternative to the opacity reset used in 3D-GS. Additionally, we extended this reduction beyond the densification phase, finding it further minimized redundancy without compromising fidelity.
Following previous methods, we trained our model for 30k iterations. The densification stops at 12000 iterations. For the version with AbsGS, $\tau=0.00067$, and $0.0016$ for the small version.
For the version with 3D-GS, $\tau=0.00028$.
We set $L=3$ and $K=3$. Our first stage lasts for 2500 iterations, the second stage for 3500 iterations, and the rest iterations are in the last stage. 
$\alpha=2^{\frac{1}{3}}$, $\lambda_s=1.6$ and $\beta=0.3$. The loss function remains the same as in 3D-GS \cite{Kerbl20233DGS}, consisting of a combination of $\mathcal{L}_1$ and D-SSIM losses. 
All experiments were done on an NVIDIA RTX3090 GPU.

\subsection{Comparisons}
\label{sec:comparison}

To assess the effectiveness of our method, we compare it with several 3D-GS variants: 3D-GS \cite{Kerbl20233DGS}, Scaffold-GS \cite{Lu2023ScaffoldGSS3}, Octree-GS \cite{Ren2024OctreeGSTC}, FreGS \cite{Zhang2024FreGS3G}, AbsGS \cite{Ye2024AbsGSRF}, Pixel-GS \cite{zhang2024pixelgs} and Mini-Splatting-D \cite{Fang2024MiniSplattingRS}. Note that Scaffold-GS and Octree-GS use a latent representation rather than the original 3D-GS representation. 
Some radiance-field-based methods are also selected for comparison, including Plenoxels \cite{Yu2021PlenoxelsRF}, Instant-NGP \cite{Mller2022InstantNG} and Mip-NeRF360 \cite{Barron2021MipNeRFAM}. 
We find that the resolution of Mip-NeRF360 dataset images varies across different works. 
Following the resolution setup in the 3D-GS work, we retrain some models on the Mip-NeRF360 dataset. 
Additionally, for works that did not report their storage usage, we report the results based on their released code.

\begin{figure*}[t!]
\includegraphics[width=\linewidth]{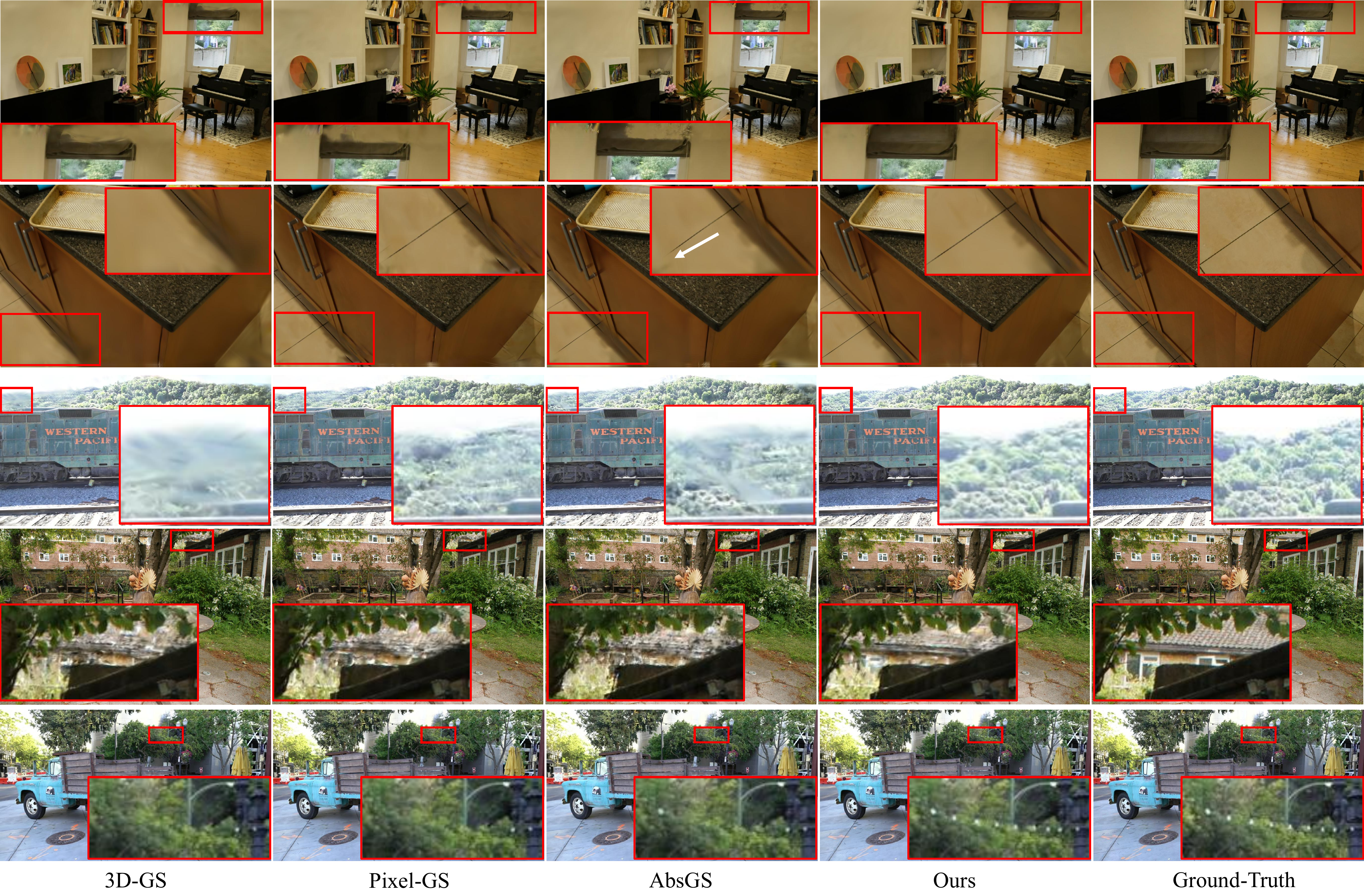}
\caption{
\textbf{Qualitative comparisons of \modelname\ with three 3D-GS variations on a variety of indoor and outdoor scenes.} 
Our approach captures more intrinsic details and acquires more complete geometry in complex scenes. 
} 
\label{fig:qualitive_results}
\end{figure*}

\paragraph{Quality Comparisons.} In Table~\ref{tab:quality}, we present quantitative results comparing the rendering quality of our method with other approaches.
From the table, we can see that our method achieves SOTA performance across all metrics on the Mip-NeRF360 \cite{Barron2021MipNeRFAM} dataset. On the Tanks\&Temples \cite{Knapitsch2017TanksAT} dataset, our method also achieves the best SSIM and LPIPS metrics, which closely reflect human perception. In the Deep Blending \cite{Hedman2018DeepBF} dataset, though our method does not achieve the highest score, it shows a notable improvement over the baseline. Additionally, our model variant without AbsGS \cite{Ye2024AbsGSRF} maintains similar performance to the full model, with only a slight reduction in LPIPS. Notably, our smaller model version achieves strong performance with reduced memory usage, showcasing the efficiency and scalability of our approach. In Fig.~\ref{fig:qualitive_results}, we present qualitative comparisons between our method and several 3D-GS variants \cite{Kerbl20233DGS, zhang2024pixelgs, Ye2024AbsGSRF} across various scenes, including both indoor and outdoor settings. It is evident that our model captures finer details, as shown in the second, third, and fourth rows, constructs missing geometry more effectively (last row), and achieves more accurate geometry reconstruction (first row).

\begin{table}[]
\centering
\caption{\textbf{Efficiency comparison with former methods.}}
\label{tab:computational_cost}
\resizebox{\linewidth}{!}{
\begin{tabular}{c|cc|cc|cc}
\toprule
Dataset & \multicolumn{2}{c|}{Mip-NeRF360} & \multicolumn{2}{c|}{Tanks\&Temples} & \multicolumn{2}{c}{Deep Blending} \\
\begin{tabular}{c|c} Method & Metrics \end{tabular}  & FPS & Time & FPS & Time & FPS & Time  \\
\midrule
\textbf{Octree-GS$^*$} ~ & 119 & 29m55s & 126 &  23m8s  & 151 & 34m38s \\ 
\textbf{Scaffold-GS$^*$} ~ & 103 & 23m53s & 137 &  15m37s  & 141 & 19m40s \\ 
\hline
\textbf{3D-GS} ~ & 103 & 27m16s & 127 &  15m34s  & 101 & 24m48s \\ 
\textbf{AbsGS} ~ & 132 & 28m48s & 193 &  13m35s  & 185 & 22m3s \\ 
\textbf{Mini-Splatting-D} ~ & 74 & 34m42s & 75 &  27m39s  & 100 & 30m37s \\ 
\textbf{Pixel-GS$^*$} ~ & 65 & 40m6s & 70 &  29m29s  & 70 & 34m1s \\ 
\hline
\textbf{Ours-Small (AbsGS)} ~ & \underline{\textbf{141}} & \underline{\textbf{20m56s}} & \underline{\textbf{206}} & \underline{\textbf{11m35s}} & \underline{\textbf{202}} & \underline{\textbf{19m26s}} \\ 
\textbf{Ours (AbsGS)} ~ & 100 & 29m24s & 150 &  16m29s  & 153 & 23m33s \\ 
\bottomrule
\end{tabular}}
\end{table}

\begin{table*}[]
\centering
\caption{\textbf{Analyzing \densifyname~upon multiple 3D-GS variants.} 
The performance consistently increases on these pipelines, indicating that our residual split is a better densification operation. 
\textbf{3D-GS$^\dagger$} indicates a retrained version of 3D-GS \cite{Kerbl20233DGS} for better performance. 
}
\label{tab:analyze_densification_method}
\resizebox{\linewidth}{!}{
\begin{tabular}{c|cccc|cccc|cccc}
\toprule
Dataset & \multicolumn{4}{c|}{Mip-NeRF360} & \multicolumn{4}{c|}{Tanks\&Temples} & \multicolumn{4}{c}{Deep Blending} \\
\begin{tabular}{c|c} Method & Metrics \end{tabular}  & PSNR \(\uparrow\) & SSIM \(\uparrow\) & LPIPS \(\downarrow\)& Mem  & PSNR \(\uparrow\) & SSIM \(\uparrow\) & LPIPS \(\downarrow\) & Mem  & PSNR \(\uparrow\) & SSIM \(\uparrow\) & LPIPS \(\downarrow\) & Mem  \\
\midrule
\textbf{3D-GS} ~ & 27.21 &\underline{\textbf{0.815}} & \underline{\textbf{0.214}} &  734MB  & 23.14 & 0.841 & 0.183 & 411MB & 29.41 & \underline{\textbf{0.903}} & 0.243 & 676MB \\ 
\textbf{3D-GS$^\dagger$} ~ & 27.38 & 0.813 & 0.215 & 821MB  & 23.66 & 0.844 & 0.179 &434MB & 29.48 & 0.900 & 0.247 &663MB \\ 
\textbf{3D-GS + \densifyname} ~ & \underline{\textbf{27.44}} & 0.813 & 0.216  & 586MB  & \underline{\textbf{23.76}} & \underline{\textbf{0.845}} & \underline{\textbf{0.178}} &318MB & \underline{\textbf{29.75}} & 0.902 & \underline{\textbf{0.240}} &546MB \\ 
\hline
\textbf{AbsGS} ~ & 27.49 & 0.820 & 0.191 & 728MB  & 23.73 & 0.853 & 0.162 & 304MB  & 29.67 & 0.902 & 0.236 & 444MB  \\ 
\textbf{AbsGS + \densifyname} ~ &\underline{\textbf{27.71}}  &\underline{\textbf{0.827}}  & \underline{\textbf{0.185}} & 712MB  & \underline{\textbf{24.05}} &\underline{\textbf{0.857}}  &\underline{\textbf{0.161}}  & 285MB  & \underline{\textbf{29.68}}  & \underline{\textbf{0.905}} &\underline{\textbf{0.234}}  & 421MB  \\ 
\hline
\textbf{Pixel-GS$^*$} ~ & 27.52  & \underline{\textbf{0.822}}  & \underline{\textbf{0.191}} & 1.32GB  & 23.78 & 0.853 & 0.151 & 1.06GB  & 28.91 & 0.892 & 0.250 & 1.09GB  \\ 
\textbf{Pixel-GS + \densifyname} ~ &\underline{\textbf{27.62}}  & 0.821  & \underline{\textbf{0.191}} & 1.00GB  & \underline{\textbf{24.05}} &\underline{\textbf{0.856}}  &\underline{\textbf{0.148}}  & 754MB &\underline{\textbf{29.75}}  & \underline{\textbf{0.899}} &\underline{\textbf{0.235}}  & 914MB  \\

\hline
\textbf{Mini-Splatting-D} ~ &  27.51 & \underline{\textbf{0.831}} & \underline{\textbf{0.176}} & 1.11GB  & 23.23 & \underline{\textbf{0.853}}  & \underline{\textbf{0.140}} & 1.01GB  &  29.88 &  \underline{\textbf{0.906}} & 0.211 & 1.09GB  \\
\textbf{Mini-Splatting-D + \densifyname} ~ &\underline{\textbf{27.64}}  & \underline{\textbf{0.831}} & \underline{\textbf{0.176}} & 1.04GB  & \underline{\textbf{23.49}} &\underline{\textbf{0.853}}  &\underline{\textbf{0.140}}  & 967MB &\underline{\textbf{29.94}}  & 0.903 &\underline{\textbf{0.210}}  & 1.02GB  \\ 
\bottomrule
\end{tabular}}
\end{table*}

\paragraph{Efficiency Comparisons.} We provide the computational costs of our method in comparison with other approaches in  Table~\ref{tab:computational_cost}. It shows that our method involves a slight increase in training time while delivering much higher rendering quality than methods with lower rendering quality like 3D-GS \cite{Kerbl20233DGS} and AbsGS \cite{Ye2024AbsGSRF}. 
Our method is also significantly more efficient than other methods like Mini-Splatting-D \cite{Fang2024MiniSplattingRS}, with comparable rendering quality. Notably, our small model provides the fastest FPS and shortest training time in all compared methods while maintaining high fidelity.

\subsection{Analysis}

\subsubsection{Residual Split}
\paragraph{Compatibility.}
We analyze the compatibility of our novel densification operation by replacing split and clone in multiple 3D-GS variants \cite{Kerbl20233DGS,Ye2024AbsGSRF,zhang2024pixelgs,Fang2024MiniSplattingRS} with \densifyname. 
The hyperparameters are set the same as in the original pipelines. 
As Table~\ref{tab:analyze_densification_method} shows, our method leads to a notable increase in PSNR metrics across all pipelines, while for other metrics, some exhibit improvements, and others remain approximately the same. Additionally, memory consumption is reduced after incorporating \densifyname~compared to the original pipeline. This underscores our method's ability to improve rendering quality and reduce redundancy, while also demonstrating its compatibility and potential for broader application across 3D-GS-oriented works as an effective alternative to split and clone.

\begin{figure}[t!]
\includegraphics[width=\linewidth]{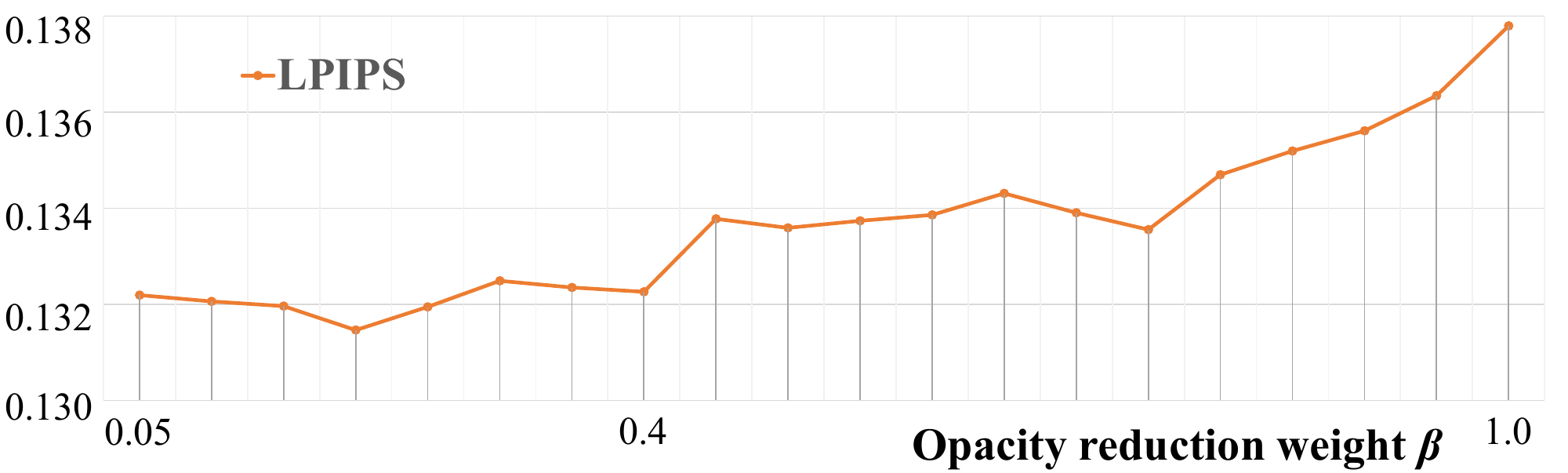}
\caption{
\textbf{Impact of the opacity reduction weight $\beta$ for \densifyname.} 
} 
\label{fig:Opacity_rw_chart}
\end{figure}

\begin{figure}[t!]
\includegraphics[width=\linewidth]{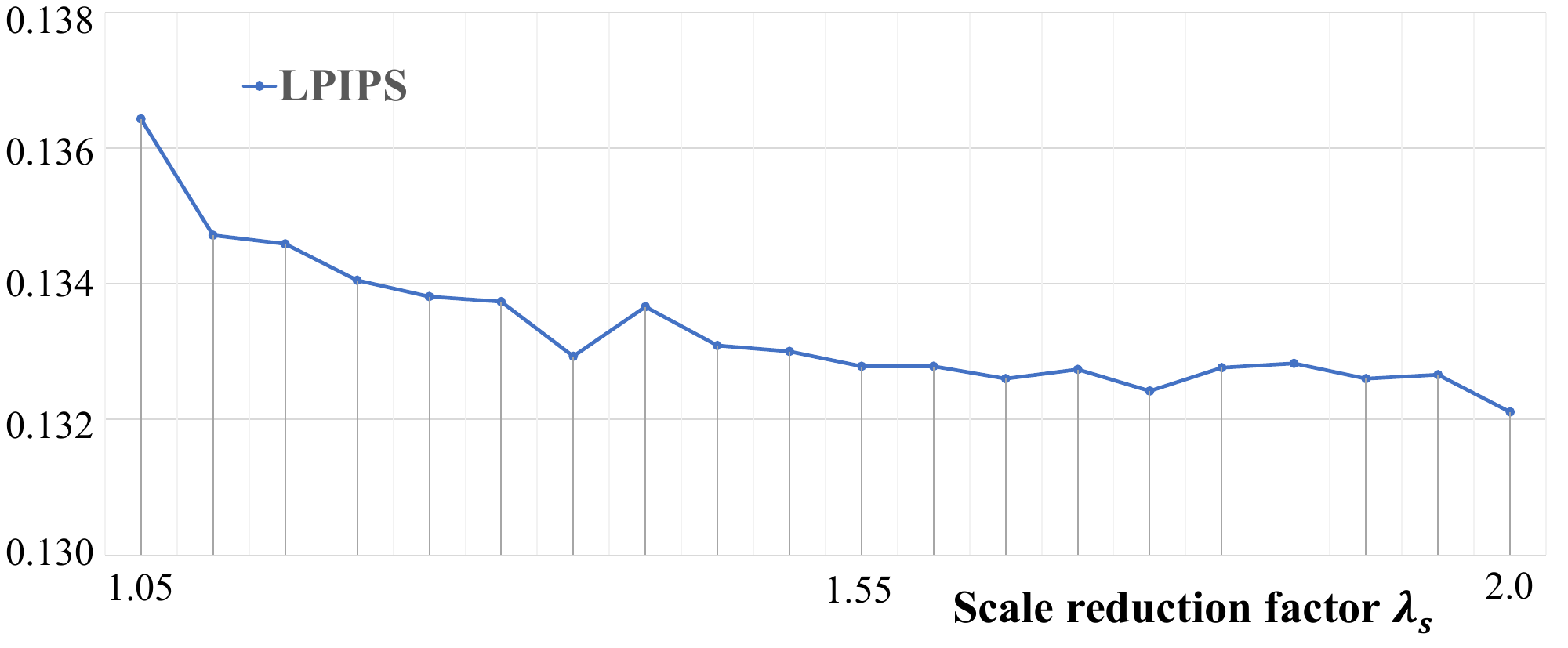}
\caption{
\textbf{Impact of the scale reduction factor $\lambda_s$ for \densifyname.} 
} 
\label{fig:Lambda_scale_chart}
\end{figure}

\paragraph{Hyperparameter Sensitivities.}
Our residual split involves some hyperparameters: opacity reduction weight $\beta$ and scale reduction factor $\lambda_s$. 
We conduct a corresponding sensitivity analysis on the Tanks\&Temples \cite{Knapitsch2017TanksAT} dataset while applying different $\beta \in [0.05,1.0)$ and $\lambda_s \in [1.05,2.0]$. 
All models are trained with similar storage. 
The LPIPS scores with different parameters are shown in Fig.~\ref{fig:Opacity_rw_chart} and \ref{fig:Lambda_scale_chart}. 
As they show, the LPIPS score remains relatively stable when $\beta<0.4$ and $\lambda_s>1.55$, indicating that our method is robust to hyperparameter variations, with only a 0.006 difference in LPIPS between the best and worst cases and a relatively broad range where rendering quality remains optimal.

\subsubsection{Ablation Studies}

\begin{table}[]
\centering
\caption{\textbf{Ablation Study on our pipeline.} 
\textbf{IP} denotes image pyramid supervision, \textbf{RS} represents using \densifyname~for densification, and \textbf{VT} indicates applying varying gradient threshold. Each component contributes to our method's effectiveness.
}
\label{tab:ablation_densification}
\resizebox{\linewidth}{!}{
\begin{tabular}{c|ccc|ccc}
\toprule
Dataset & \multicolumn{3}{c}{Mip-NeRF360} & \multicolumn{3}{c}{Deep Blending} \\
\begin{tabular}{c|c} Method & Metrics \end{tabular}  & PSNR \(\uparrow\) & SSIM \(\uparrow\) & LPIPS \(\downarrow\) & PSNR \(\uparrow\) & SSIM \(\uparrow\) & LPIPS \(\downarrow\)\\
\midrule
\textbf{Base} ~ & 27.41 & 0.817 & 0.189 & 29.60 & \cellcolor{tabthird}0.899 & 0.240\\
\textbf{Base + RS} ~ & \cellcolor{tabthird}27.66 & \cellcolor{tabthird}0.825 & 0.183 & \cellcolor{tabthird}29.68 & \cellcolor{tabsecond}0.900 & \cellcolor{tabthird}0.233\\ 
\textbf{Base + IP} ~ & 27.54 & 0.823 & \cellcolor{tabthird}0.182 & 29.05 & 0.896 & 0.245\\ 
\textbf{Base + RS + IP} ~ & \cellcolor{tabsecond}27.88 & \cellcolor{tabsecond}0.831 & \cellcolor{tabsecond}0.178 & \cellcolor{tabsecond}29.82 & \cellcolor{tabfirst}0.902 & \cellcolor{tabsecond}0.230 \\ 
\hline
\textbf{Base + RS + IP + VT (full)} &\cellcolor{tabfirst}28.00 & \cellcolor{tabfirst}0.833 & \cellcolor{tabfirst}0.174 & \cellcolor{tabfirst}29.91 & \cellcolor{tabfirst}0.902 & \cellcolor{tabfirst}0.227 \\
\bottomrule
\end{tabular}}
\end{table}

To assess the effectiveness of our pipeline's key modules, including \densifyname, image pyramid as supervision, and the varying gradient threshold, we conduct ablation studies on the Mip-NeRF360 \cite{Barron2021MipNeRFAM} and Deep Blending \cite{Hedman2018DeepBF} datasets. 
For the varying gradient threshold, we only conduct studies on its removal, as it is highly interdependent with the other two modules. 
As the results in Table~\ref{tab:ablation_densification} shows, both \densifyname~and image pyramid supervision individually boost the baseline's performance on the Mip-NeRF360 dataset. 

However, on Deep Blending, we observe a performance drop in \textbf{Base+IP}. This is because the dataset contains many weakly textured areas. As noted in Sec.~\ref{sec:intro}, split and clone tend to fill such regions with an excessive number of small Gaussians. This makes them prone to overfitting the training views and lead to reduced performance on test views. 
The issue is amplified in the early iterations of a coarse-to-fine schedule, where only coarse scene information is available, making weakly textured areas even harder to resolve. As a result, these small Gaussians struggle to construct accurate geometry, leading to the performance drop observed in \textbf{Base+IP}. 

Conversely, when using \textbf{RS}, weakly textured areas in early iterations are represented by only a few Gaussians, reducing overfitting. In later iterations, as finer information becomes available, \textbf{RS} progressively introduces more small Gaussians. As a result, \textbf{RS} benefits from progressive training, as evidenced by the improved performance of \textbf{Base+RS+IP} over \textbf{Base+RS}. 
This further proves \textbf{RS} is more effective than split and clone, especially in a coarse-to-fine schedule. 
Additionally, applying the varying gradient threshold (our full model) further boosts performance on both datasets, validating its effectiveness. 

\subsection{Discussions and Limitations}
\label{sec:limitation}
The above results demonstrate that our method significantly enhances rendering quality on the Mip-NeRF360 \cite{Barron2021MipNeRFAM} and Tanks\&Temples \cite{Knapitsch2017TanksAT} datasets, achieving state-of-the-art performance. However, for the Deep Blending dataset \cite{Hedman2018DeepBF}, our method fails to achieve the best performance on every metric. While our approach implicitly constrains the number of Gaussians in weakly textured regions, which helps mitigate overfitting and improves quality, it does not include explicit regularization. This suggests an opportunity for further optimization in future work.


%% file: sec/5_conclusion.tex
\section{Conclusion}

In this paper, we analyze the limitations of the current 3D-GS densification operation. To address these issues, we propose \densifyname, a residual empowered densification operation that adaptively handles both over-reconstruction and under-reconstruction. Additionally, we introduce \modelname, a coarse-to-fine 3D-GS pipeline to supplement \densifyname. 
Specifically, we use an image pyramid for supervision to gradually introduce fine details during Gaussian optimization. 
We also design a Gaussian selection scheme to further densify coarse Gaussians progressively. 
Our pipeline achieves SOTA rendering quality on several datasets while our \densifyname~is highly compatible with other 3D-GS variants.



%% file: sec/6_ack.tex
\section*{Acknowledgements}
This work was supported by CAS Project for Young Scientists in Basic Research (YSBR-116).

%% file: sec/X_suppl.tex
\clearpage
\setcounter{page}{1}
\maketitlesupplementary

\section{Additional Experiment Results}

In this section, we provide additional experimental results of our work. 

\subsection{Evaluation on large-scale urban datasets} 

To evaluate the performance of our approach on large-scale urban datasets, we performed experiments using the Waymo \cite{Sun2019ScalabilityIP} dataset and compared our results against GaussianPro \cite{Cheng2024GaussianPro3G}. The results are shown in the table below. It can be seen that our method achieves better results across all metrics. 

\begin{table}[h]
\small
\centering
\caption{\textbf{Results on the Waymo dataset compared with GaussianPro.} }
\label{tab:db_stddev}
\resizebox{\linewidth}{!}{
\begin{tabular}{c|cccc}
\toprule
Dataset & \multicolumn{4}{c}{Waymo}  \\
\begin{tabular}{c|c} Method & Metrics \end{tabular}  & PSNR \(\uparrow\) & SSIM \(\uparrow\) & LPIPS \(\downarrow\) & Mem. \\
\midrule
\textbf{GaussianPro} & 34.68 & 0.949 & \underline{\textbf{0.191}} & 297MB 
\\

\textbf{Ours} & \underline{\textbf{35.29}} & \underline{\textbf{0.951}} & \underline{\textbf{0.191}} & 284MB  \\
\bottomrule
\end{tabular}}
\end{table}

\subsection{Variance of PSNR on Deep Blending Dataset} 
A notable variance in the PSNR metrics of our method is observed on the Deep Blending \cite{Hedman2018DeepBF} dataset. We tested our method on the Deep Blending dataset 10 times and reported the average performance and population standard deviation. The results are shown in Table~\ref{tab:db_stddev}. The table demonstrates that the PSNR scores on the Deep Blending dataset exhibit a notable variance. 
The underlying reason, as discussed in our limitations, is that our method does not address issues related to occlusion and overfitting, making the Deep Blending dataset particularly challenging for our approach. This highlights opportunities for future optimization and improvement. Furthermore, we report the average performance and population standard deviation over 10 runs for the other two datasets, Mip-NeRF360 \cite{Barron2021MipNeRFAM} and Tanks\&Temples \cite{Knapitsch2017TanksAT}, as presented in Table~\ref{tab:mip_stddev} and Table~\ref{tab:tt_stddev}, respectively. The PSNR metrics for these two datasets do not show a notable variance, showcasing the robustness of our method in most scenarios.

\begin{table}[H]
\centering
\caption{\textbf{Average and standard deviation results of three metrics on the Deep Blending dataset.} }
\label{tab:db_stddev}
\resizebox{\linewidth}{!}{
\begin{tabular}{c|cc|cc|cc}
\toprule
Dataset & \multicolumn{6}{c}{Deep Blending}\\

\begin{tabular}{c|c} Method & Metrics \end{tabular}  & \multicolumn{2}{c|}{PSNR $\uparrow$} & \multicolumn{2}{c|}{SSIM $\uparrow$} & \multicolumn{2}{c}{LPIPS $\downarrow$} \\
  & Avg & Stdev & Avg & Stdev & Avg & Stdev \\
\midrule
\textbf{Ours-Small (AbsGS)} ~ & 29.71 & 0.211 & 0.903 & 0.002 & 0.235 & 0.002\\ 
\textbf{Ours (3D-GS)} ~ & 29.64 & 0.205 & 0.900 & 0.003 & 0.233 & 0.002\\ 
\textbf{Ours (AbsGS)} ~ & 29.68 & 0.173 & 0.900 & 0.002 & 0.228 & 0.002\\ 
\bottomrule
\end{tabular}}
\end{table}

\begin{table}[H]
\centering
\caption{\textbf{Average and standard deviation results of three metrics on the Mip-NeRF360 dataset.} }
\label{tab:mip_stddev}
\resizebox{\linewidth}{!}{
\begin{tabular}{c|cc|cc|cc}
\toprule
Dataset & \multicolumn{6}{c}{Mip-NeRF360}\\

\begin{tabular}{c|c} Method & Metrics \end{tabular}  & \multicolumn{2}{c|}{PSNR $\uparrow$} & \multicolumn{2}{c|}{SSIM $\uparrow$} & \multicolumn{2}{c}{LPIPS $\downarrow$} \\
  & Avg & Stdev & Avg & Stdev & Avg & Stdev \\
\midrule
\textbf{Ours-Small (AbsGS)} ~ & 27.93 & 0.010 & 0.830 & 0.001 & 0.191 & 0.001\\ 
\textbf{Ours (3D-GS)} ~ & 27.99 & 0.011 & 0.831 & 0.001 & 0.187 & 0.001\\ 
\textbf{Ours (AbsGS)} ~ & 27.99 & 0.011 & 0.833 & 0.001 & 0.174 & 0.001\\ 
\bottomrule
\end{tabular}}
\end{table}

\begin{table}[H]
\centering
\caption{\textbf{Average and standard deviation results of three metrics on the Tank\&Temples dataset.} }
\label{tab:tt_stddev}
\resizebox{\linewidth}{!}{
\begin{tabular}{c|cc|cc|cc}
\toprule
Dataset & \multicolumn{6}{c}{Tank\&Temples}\\

\begin{tabular}{c|c} Method & Metrics \end{tabular}  & \multicolumn{2}{c|}{PSNR $\uparrow$} & \multicolumn{2}{c|}{SSIM $\uparrow$} & \multicolumn{2}{c}{LPIPS $\downarrow$} \\
  & Avg & Stdev & Avg & Stdev & Avg & Stdev \\
\midrule
\textbf{Ours-Small (AbsGS)} ~ & 24.21 & 0.081 & 0.862 & 0.001 & 0.151 & 0.001\\ 
\textbf{Ours (3D-GS)} ~ & 24.26 & 0.058 & 0.864 & 0.001 & 0.141 & 0.001\\ 
\textbf{Ours (AbsGS)} ~ & 24.27 & 0.063 & 0.866 & 0.001 & 0.133 & 0.001\\ 
\bottomrule
\end{tabular}}
\end{table}

\subsection{Further Analysis of Residual Split on 3D-GS}

\begin{table*}[]
\centering
\caption{\textbf{Results of \densifyname~upon 3D-GS with varied gradient threshold.}}
\label{tab:3dgs_rs_vg}
\resizebox{\linewidth}{!}{
\begin{tabular}{c|cccc|cccc|cccc}
\toprule
Dataset & \multicolumn{4}{c|}{Mip-NeRF360} & \multicolumn{4}{c|}{Tanks\&Temples} & \multicolumn{4}{c}{Deep Blending} \\
\begin{tabular}{c|c} Method & Metrics \end{tabular}  & PSNR \(\uparrow\) & SSIM \(\uparrow\) & LPIPS \(\downarrow\)& Mem  & PSNR \(\uparrow\) & SSIM \(\uparrow\) & LPIPS \(\downarrow\) & Mem  & PSNR \(\uparrow\) & SSIM \(\uparrow\) & LPIPS \(\downarrow\) & Mem  \\
\midrule
\textbf{3D-GS} ~ & 27.21 &\underline{\textbf{0.815}} & 0.214 &  734MB  & 23.14 & 0.841 & 0.183 & 411MB & 29.41 & \underline{\textbf{0.903}} & 0.243 & 676MB \\ 
\textbf{3D-GS + \densifyname} ~ & \underline{\textbf{27.48}} & \underline{\textbf{0.815}} & \underline{\textbf{0.213}}  & 722MB  & \underline{\textbf{23.85}} & \underline{\textbf{0.846}} & \underline{\textbf{0.174}} &374MB & \underline{\textbf{29.83}} & 0.901 & \underline{\textbf{0.238}} &640MB \\ 
\bottomrule
\end{tabular}}
\end{table*}

In our paper, we tested the compatibility of our \densifyname~across multiple pipelines without modifying any hyper-parameters. Here, we present the results of applying \densifyname~to 3D-GS \cite{Kerbl20233DGS} while adjusting the densification threshold $\tau$ to maintain the same storage size as 3D-GS, further demonstrating the capability of our method to enhance rendering quality. The results are shown in Table \ref{tab:3dgs_rs_vg}, showing that our method yields a further performance increase.

\begin{table*}[]
\centering
\caption{\textbf{Results of our method without extended periodic opacity reduction.} Here, \textbf{No EPOR} refers to our method trained without applying the extended periodic opacity reduction. \textbf{No EPOR + CT} refers to the model without the extended periodic opacity reduction but with an adjusted gradient threshold to ensure that the final storage size is approximately the same as our full model with extended periodic opacity reduction.}
\label{tab:no_periodic_opacity_reduction}
\resizebox{\linewidth}{!}{
\begin{tabular}{c|cccc|cccc|cccc}
\toprule
Dataset & \multicolumn{4}{c|}{Mip-NeRF360} & \multicolumn{4}{c|}{Tanks\&Temples} & \multicolumn{4}{c}{Deep Blending} \\
\begin{tabular}{c|c} Method & Metrics \end{tabular}  & PSNR \(\uparrow\) & SSIM \(\uparrow\) & LPIPS \(\downarrow\)& Mem  & PSNR \(\uparrow\) & SSIM \(\uparrow\) & LPIPS \(\downarrow\) & Mem  & PSNR \(\uparrow\) & SSIM \(\uparrow\) & LPIPS \(\downarrow\) & Mem  \\
\midrule
\textbf{Ours (No EPOR)} & \underline{\textbf{28.02}}& \underline{\textbf{0.833}} & \underline{\textbf{0.173}} & 973MB  & 24.32 & \underline{\textbf{0.867}} & \underline{\textbf{0.130}} & 591MB  & 29.86 & 0.902 & \underline{\textbf{0.227}} & 911MB \\
\textbf{Ours (No EPOR + CT)} & 28.00 & \underline{\textbf{0.833}} & 0.181 & 608MB  & 24.35 & 0.866 & 0.139 & 380MB  & \underline{\textbf{29.96}} & \underline{\textbf{0.904}} & 0.230 & 564MB \\
\textbf{Ours (full)} & 28.00 & \underline{\textbf{0.833}} & 0.174 & 698MB  & \underline{\textbf{24.38}} & \underline{\textbf{0.867}} & 0.132 & 351MB  & 29.91 & 0.902 & \underline{\textbf{0.227}} & 586MB  \\
\bottomrule
\end{tabular}}
\end{table*}

\subsection{Results using random initialization} 

To assess the robustness of our method to random initialization, we conducted experiments using randomly initialized point clouds and compared our approach with other methods trained on the same random point clouds. The results are shown in Table \ref{tab:random_results}. For the Mip-NeRF360 \cite{Barron2021MipNeRFAM} and Tanks\&Temples \cite{Knapitsch2017TanksAT} datasets, our model shows a noticeable drop in quality compared to when initialized with SfM \cite{schoenberger2016sfm}. However, compared to other methods, it still achieves the best performance across all three metrics on Mip-NeRF360, and the highest SSIM and LPIPS on Tanks\&Temples. As for Deep Blending \cite{Hedman2018DeepBF}, our method and others, excluding Scaffold-GS and Octree-GS, do not exhibit a significant performance drop. While we do not achieve the highest scores, our method remains competitive, ranking second across all metrics. 
\begin{table*}[h!]
\small
\centering
\caption{\textbf{Results using random initialized point clouds, compared with other methods.}}
\label{tab:random_results}
\resizebox{\linewidth}{!}{
\begin{tabular}{c|cccc|cccc|cccc}
\toprule
Dataset & \multicolumn{4}{c|}{Mip-NeRF360} & \multicolumn{4}{c|}{Tanks\&Temples} & \multicolumn{4}{c}{Deep Blending} \\
\begin{tabular}{c|c} Method & Metrics \end{tabular}  & PSNR \(\uparrow\) & SSIM \(\uparrow\) & LPIPS \(\downarrow\) & Mem. & PSNR \(\uparrow\) & SSIM \(\uparrow\) & LPIPS \(\downarrow\) & Mem. & PSNR \(\uparrow\) & SSIM \(\uparrow\) & LPIPS \(\downarrow\) & Mem. \\
\midrule
\textbf{Scaffold-GS} & \cellcolor{tabthird}25.90 & 0.743 & 0.301 & 157MB & \cellcolor{tabthird}22.00 & 0.768 & 0.282 & 51MB & 29.48 & 0.891 & 0.285 & 56MB \\ 
\textbf{Octree-GS} & 25.70 & 0.739 & 0.292 & 190MB & \cellcolor{tabfirst}22.15 & \cellcolor{tabsecond}0.794 & 0.226 & 81MB & 28.29 & 0.873 &0.300 & 92MB\\ 
\hline
\textbf{3D-GS} & \cellcolor{tabthird}25.90 & 0.769 & 0.266 & 632MB & 21.19 & 0.777 & 0.241 & 339MB & 29.42 & \cellcolor{tabthird}0.897 &0.253 & 605MB \\ 
\textbf{AbsGS} & 25.50 & \cellcolor{tabthird}0.773 & 0.226 & 741MB & 20.89 & 0.786 & \cellcolor{tabthird}0.211 & 304MB & \cellcolor{tabthird}29.52 & \cellcolor{tabthird}0.897 &\cellcolor{tabthird}0.241 & 478MB \\ 
\textbf{Mini-Splatting-D} & 25.18 & \cellcolor{tabsecond}0.792 & \cellcolor{tabsecond}0.207 & 1.10GB & 19.77 & 0.760 & 0.220 & 1.02GB &\cellcolor{tabfirst}29.76 & \cellcolor{tabfirst}0.901 & \cellcolor{tabfirst}0.212 & 1.09GB \\
\textbf{Pixel-GS} & \cellcolor{tabsecond}26.23 & \cellcolor{tabsecond}0.792  & \cellcolor{tabthird}0.221 & 1.14GB & 21.10 & \cellcolor{tabthird}0.788 & \cellcolor{tabsecond}0.207 & 939MB & 28.76 & 0.889 & 0.256 & 1.04GB \\
\hline
\textbf{Ours-Random} & \cellcolor{tabfirst}26.38 & \cellcolor{tabfirst}0.799 & \cellcolor{tabfirst}0.200 & 750MB & \cellcolor{tabsecond}22.02 & \cellcolor{tabfirst}0.815 & \cellcolor{tabfirst}0.178  & 282MB & \cellcolor{tabsecond}29.57 & \cellcolor{tabsecond}0.899 &\cellcolor{tabsecond}0.234 & 464MB \\
\hline
\textbf{Ours-SfM} & 28.00 & 0.833 & 0.174 & 698MB  & 24.38 & 0.867 &0.132 & 351MB  & 29.91 & 0.902 & 0.227 & 586MB  \\
\bottomrule
\end{tabular}}
\end{table*}

\subsection{Periodic Opacity Reduction} 
3D-GS \cite{Kerbl20233DGS} often encounters redundancy issues, which we categorize into two main causes: the first is the use of an excessive number of small-scale Gaussians to represent coarse areas, and the second is the overlap between Gaussians, resulting in redundant Gaussians that contribute minimally to the rendered image. Our \densifyname~can solve the first issue but does not address the second. To tackle the second issue, as mentioned in our experiment settings, we implemented the periodic opacity reduction technique in \cite{Ye2024AbsGSRF, Bul2024RevisingDI}. Specifically, the opacity of Gaussians is periodically reduced, causing the opacity of those that contribute little to the rendered image to diminish to a negligible value, allowing them to be effectively pruned. Additionally, we extended this operation to occur after the densification stage, discovering that it further reduces redundancy without compromising fidelity. Here, we present the experimental results of our method without extending the periodic opacity reduction operation, as shown in Table~\ref{tab:no_periodic_opacity_reduction}. From the table, we can observe that extending the periodic opacity reduction operation effectively reduces redundancy without compromising the fidelity of the rendered images. Furthermore, when the operation is not extended and the gradient threshold is adjusted to maintain storage approximately equal to our full model, the rendering quality shows only a slight decrease in LPIPS metrics. This indicates that while extending the operation can enhance the performance of our method, its overall impact on the performance improvement is relatively minor.

\subsection{Image Resolution of Mip-NeRF360 Dataset} 
In our paper, we mentioned that the image resolution on the Mip-NeRF360 \cite{Barron2021MipNeRFAM} dataset varies across different works. 3D-GS \cite{Kerbl20233DGS} uses the provided ``images\_4", the official downscaled 4 times images from the dataset, for outdoor scenes and ``images\_2" for indoor scenes. Most works \cite{Ye2024AbsGSRF, Fang2024MiniSplattingRS, Zhang2024FreGS3G} follow the 3D-GS setting, but Scaffold-GS \cite{Lu2023ScaffoldGSS3}, Octree-GS \cite{Ren2024OctreeGSTC}, Pixel-GS \cite{zhang2024pixelgs} uses different settings. 
Specifically, Scaffold-GS and Octree-GS downscale original images to 1.6K resolution, while Pixel-GS downscales original images by a factor of 4 for outdoor scenes and 2 for indoor scenes during the training process, instead of using the provided downscaled images. This discrepancy results in differences in experimental settings. To further demonstrate the performance of our method, we provide its results under the two different settings, as shown in Table \ref{tab:mip_scoc} and \ref{tab:mip_pixel}. The tables show that our method prevails even under these alternate settings.

\begin{table}[H]
\centering
\caption{\textbf{Results on the Mip-NeRF360 dataset with same setting as Scaffold-GS and Octree-GS.} }
\label{tab:mip_scoc}
\resizebox{\linewidth}{!}{
\begin{tabular}{c|ccc}
\toprule
Dataset & \multicolumn{3}{c}{Mip-NeRF360} \\
\begin{tabular}{c|c} Method & Metrics \end{tabular}  & PSNR \(\uparrow\) & SSIM \(\uparrow\) & LPIPS \(\downarrow\)\\
\midrule
\textbf{Scaffold-GS } ~ &  27.71 & 0.813 & 0.221\\
\textbf{Octree-GS } ~ &  27.73 & 0.815& 0.217 \\ 

\hline
\textbf{Ours-Small (AbsGS)} ~ & \cellcolor{tabthird}28.04 & \cellcolor{tabthird}0.831 & \cellcolor{tabthird}0.191\\
\textbf{Ours (3D-GS)} ~ & \cellcolor{tabsecond}28.10 & \cellcolor{tabsecond}0.832 & \cellcolor{tabsecond}0.187\\
\textbf{Ours (AbsGS)} ~ & \cellcolor{tabfirst}28.14 & \cellcolor{tabfirst}0.834 & \cellcolor{tabfirst}0.174\\ 
\bottomrule
\end{tabular}}
\end{table}

\begin{table}[]
\centering
\caption{\textbf{Results on the Mip-NeRF360 dataset with same setting as Pixel-GS.} }
\label{tab:mip_pixel}
\resizebox{\linewidth}{!}{
\begin{tabular}{c|ccc}
\toprule
Dataset & \multicolumn{3}{c}{Mip-NeRF360} \\
\begin{tabular}{c|c} Method & Metrics \end{tabular}  & PSNR \(\uparrow\) & SSIM \(\uparrow\) & LPIPS \(\downarrow\)\\
\midrule
\textbf{Pixel-GS } ~ & 27.88 & 0.834 & \cellcolor{tabthird}0.176\\

\hline
\textbf{Ours-Small (AbsGS)} ~ & \cellcolor{tabthird}28.21 & \cellcolor{tabthird}0.841 & \cellcolor{tabthird}0.176\\
\textbf{Ours (3D-GS)} ~ & \cellcolor{tabsecond}28.28 & \cellcolor{tabsecond}0.843 & \cellcolor{tabsecond}0.172\\
\textbf{Ours (AbsGS)} ~ & \cellcolor{tabfirst}28.30 & \cellcolor{tabfirst}0.845 & \cellcolor{tabfirst}0.160\\ 
\bottomrule
\end{tabular}}
\end{table}

\subsection{Per Scene Results} We present the per-scene results of the used metrics in Table \ref{tab:ttdb_psnr}-\ref{tab:mip_lpips}. For works that did not report per-scene results, we obtained and reported them using their released code. The works selected for comparison are mainly the same as in the paper:  3D-GS \cite{Kerbl20233DGS}, Scaffold-GS \cite{Lu2023ScaffoldGSS3}, Octree-GS \cite{Ren2024OctreeGSTC}, AbsGS \cite{Ye2024AbsGSRF}, Pixel-GS \cite{zhang2024pixelgs}, Mini-Splatting-D \cite{Fang2024MiniSplattingRS}, Plenoxels \cite{Yu2021PlenoxelsRF}, Instant-NGP \cite{Mller2022InstantNG} and Mip-NeRF360 \cite{Barron2021MipNeRFAM}. Note that we did not show the results of FreGS 
\cite{Zhang2024FreGS3G} since they did not report their per-scene results or release their code. Works retrained on the Mip-NeRF360 \cite{Barron2021MipNeRFAM} are marked with a $^*$.

\begin{table}[H]
\centering
\caption{\textbf{Per-scene PSNR metrics of Tank\&Temples and Deep Blending dataset.} 
}
\label{tab:ttdb_psnr}
\resizebox{\linewidth}{!}{
\begin{tabular}{c|cc|cc}
\toprule
\begin{tabular}{c|c} Method & Scenes \end{tabular}  & Truck & Train & Dr Johnson & Playroom\\
\midrule
\textbf{Plenoxels} & 23.22 & 18.93 & 23.14 & 22.98 \\ 

\textbf{Instant-NGP} & 23.38  & 20.46 & 28.26 & 21.67 \\ 

\textbf{Mip-NeRF360} & 24.91 & 19.52 & 29.14 & 29.66\\
\hline
\textbf{Scaffold-GS} &25.77 & 22.15 & \cellcolor{tabsecond}29.80 & \cellcolor{tabsecond}30.62\\
\textbf{Octree-GS} & \cellcolor{tabfirst}26.27& \cellcolor{tabfirst}22.77 & \cellcolor{tabfirst}29.87&\cellcolor{tabfirst}30.95 \\

\hline
\textbf{3D-GS} &25.19 & 21.10 & 28.77 & 30.04
 \\ 
\textbf{AbsGS} & 25.74 & 21.72 & 29.20 & 30.14 \\ 

\textbf{Pixel-GS} &25.49 & 22.13 & 28.02 & 29.79\\
\textbf{Mini-Splatting-D} & 25.43 & 21.04 & 29.32 & \cellcolor{tabthird}30.43 \\
\hline
\textbf{Ours-Small (AbsGS)} & 26.07& 22.40& \cellcolor{tabthird}29.64& 30.38 \\
\textbf{Ours (3D-GS)} & \cellcolor{tabthird}26.08& \cellcolor{tabsecond}22.65 & 29.54& 30.32 \\
\textbf{Ours (AbsGS)} & \cellcolor{tabsecond}26.14& \cellcolor{tabthird}22.61& 29.51& 30.32\\
\bottomrule
\end{tabular}}
\end{table}

\begin{table}[H]
\centering
\caption{\textbf{Per-scene SSIM metrics of Tank\&Temples and Deep Blending dataset.} 
}
\label{tab:ttdb_ssim}
\resizebox{\linewidth}{!}{
\begin{tabular}{c|cc|cc}
\toprule
\begin{tabular}{c|c} Method & Scenes \end{tabular}  & Truck & Train & Dr Johnson & Playroom\\
\midrule
\textbf{Plenoxels} & 0.774 & 0.663 & 0.787 & 0.802\\ 

\textbf{Instant-NGP} & 0.800 & 0.689 & 0.854 & 0.779 \\ 

\textbf{Mip-NeRF360} & 0.857 & 0.660 & 0.901 & 0.900 \\ 
\hline
\textbf{Scaffold-GS} & 0.883 & 0.822 & \cellcolor{tabsecond}0.907 & 0.904
\\
\textbf{Octree-GS} &  \cellcolor{tabfirst}0.896 &  \cellcolor{tabthird}0.835 &  \cellcolor{tabfirst}0.912&\cellcolor{tabfirst}0.914 \\

\hline
\textbf{3D-GS} & 0.879 & 0.802 & 0.899 & 0.906 \\

\textbf{AbsGS} & 0.888 & 0.818 & 0.898 & \cellcolor{tabthird}0.907 \\

\textbf{Pixel-GS} & 0.883 & 0.823 & 0.885 & 0.899\\
\textbf{Mini-Splatting-D} & \cellcolor{tabthird}0.890 & 0.817 & \cellcolor{tabthird}0.905 & \cellcolor{tabsecond}0.908 \\
\hline
\textbf{Ours-Small (AbsGS)} & 0.889& 0.834& \cellcolor{tabthird}0.905& 0.907 \\
\textbf{Ours (3D-GS)}& \cellcolor{tabthird}0.890& \cellcolor{tabsecond}0.840 & 0.901& 0.905 \\
\textbf{Ours (AbsGS)} & \cellcolor{tabsecond}0.893& \cellcolor{tabfirst}0.841& 0.900& 0.905\\
\bottomrule
\end{tabular}}
\end{table}

\begin{table}[H]
\centering
\caption{\textbf{Per-scene LPIPS metrics of Tank\&Temples and Deep Blending dataset.} 
}
\label{tab:ttdb_lpips}
\resizebox{\linewidth}{!}{
\begin{tabular}{c|cc|cc}
\toprule
\begin{tabular}{c|c} Method & Scenes \end{tabular}  & Truck & Train & Dr Johnson & Playroom\\
\midrule
\textbf{Plenoxels} & 0.335 & 0.422 & 0.521 & 0.499\\ 

\textbf{Instant-NGP} & 0.249 & 0.360 & 0.352 & 0.428 \\ 

\textbf{Mip-NeRF360} & 0.159 & 0.354 & 0.237 & 0.252 \\ 
\hline
\textbf{Scaffold-GS} &0.147 & 0.206 & 0.250 & 0.258 \\
\textbf{Octree-GS} & \cellcolor{tabthird}0.118 & 0.198 &   \cellcolor{tabsecond}0.231&  0.244 \\

\hline
\textbf{3D-GS} & 0.148 & 0.218 & 0.244 & 0.241 \\

\textbf{AbsGS} & 0.132 & 0.193 & 0.241 & 0.233 \\

\textbf{Pixel-GS} &  0.122 & 0.180 & 0.258 &  0.243\\
\textbf{Mini-Splatting-D} & \cellcolor{tabfirst}0.100 & 0.181 & \cellcolor{tabfirst}0.218 & \cellcolor{tabfirst}0.204 \\
\hline
\textbf{Ours-Small (AbsGS)} & 0.124& \cellcolor{tabthird}0.177& 0.233& 0.235 \\
\textbf{Ours (3D-GS)}& \cellcolor{tabthird}0.118& \cellcolor{tabsecond}0.164 & \cellcolor{tabthird}0.232& \cellcolor{tabthird}0.231 \\
\textbf{Ours (AbsGS)} & \cellcolor{tabsecond}0.106& \cellcolor{tabfirst}0.159& \cellcolor{tabsecond}0.231& \cellcolor{tabsecond}0.223\\
\bottomrule
\end{tabular}}
\end{table}

\begin{table*}[]
\centering
\caption{\textbf{Per-scene PSNR metrics of Mip-NeRF360 dataset.} 
}
\label{tab:mip_psnr}
\resizebox{0.86\linewidth}{!}{
\begin{tabular}{c|ccccc|cccc}
\toprule
\begin{tabular}{c|c} Method & Scenes \end{tabular}  & bicycle & flowers & garden & stump & treehill  & room & counter & kitchen & bonsai\\
\midrule
\textbf{Plenoxels} & 21.91 & 20.10 & 23.49 & 20.66 & 22.25 & 27.59 & 23.62 & 23.42 & 24.67 \\ 

\textbf{Instant-NGP} & 22.17 & 20.65 & 25.07 & 23.47 & 22.37  & 29.69 & 26.69 & 29.48 & 30.69 \\ 

\textbf{Mip-NeRF360} & 24.31 & 21.65 & 26.88 & 26.18 & 22.93 & 31.47 & 29.45 & 31.99 & \cellcolor{tabfirst}33.40 \\
\hline
\textbf{Scaffold-GS$^*$} & 25.07& 21.36 &27.35&26.64&\cellcolor{tabfirst}22.93&32.04&\cellcolor{tabfirst}29.50&31.63&32.71 \\
\textbf{Octree-GS$^*$} & 25.05& 21.33 &27.58&26.41&22.81&32.09&\cellcolor{tabsecond}29.48&30.89&31.71 \\

\hline
\textbf{3D-GS} & 25.25 & 21.52 & 27.41 & 26.55 & 22.49 & 30.63 & 28.70 & 30.32 & 31.98 \\ 
\textbf{AbsGS} &  \cellcolor{tabthird}25.29 &21.35& 27.48& 26.71 &21.99&31.61& 29.03& 31.62 &32.32 \\ 

\textbf{Pixel-GS$^*$} & 25.21& 21.49& 27.42& 26.84& 22.09& 31.45& 29.05& 31.65& 32.53 \\
\textbf{Mini-Splatting-D} & \cellcolor{tabsecond}25.55 & 21.50 & 27.67 & 27.11 & 22.13 & 31.41 & 28.72 & 31.75 & 31.72 \\
\hline
\textbf{Ours-Small (AbsGS)} & \cellcolor{tabfirst}25.62& \cellcolor{tabsecond}21.80& \cellcolor{tabsecond}27.72& \cellcolor{tabfirst}27.27& \cellcolor{tabsecond}22.90& \cellcolor{tabthird}32.41& 29.32& \cellcolor{tabthird}31.95& 32.48 \\
\textbf{Ours (3D-GS)} & \cellcolor{tabsecond}25.55& \cellcolor{tabfirst}22.07& \cellcolor{tabthird}27.71& \cellcolor{tabthird}27.12& \cellcolor{tabthird}22.82& \cellcolor{tabsecond}32.48& \cellcolor{tabthird}29.41& \cellcolor{tabsecond}32.09& \cellcolor{tabthird}32.74 \\
\textbf{Ours (AbsGS)} & \cellcolor{tabfirst}25.62& \cellcolor{tabthird}21.78& \cellcolor{tabfirst}27.82& \cellcolor{tabsecond}27.19& 22.57& \cellcolor{tabfirst}32.51& \cellcolor{tabfirst}29.50& \cellcolor{tabfirst}32.26& \cellcolor{tabsecond}32.78 \\
\bottomrule
\end{tabular}}
\end{table*}

\begin{table*}[]
\centering
\caption{\textbf{Per-scene SSIM metrics of Mip-NeRF360 dataset.} 
}
\label{tab:mip_ssim}
\resizebox{0.86\linewidth}{!}{
\begin{tabular}{c|ccccc|cccc}
\toprule
\begin{tabular}{c|c} Method & Scenes \end{tabular}  & bicycle & flowers & garden & stump & treehill  & room & counter & kitchen & bonsai\\
\midrule
\textbf{Plenoxels} & 0.496 & 0.431 & 0.6063 & 0.523 & 0.509 & 0.8417 & 0.759 & 0.648 & 0.814 \\ 

\textbf{Instant-NGP} & 0.512 & 0.486 & 0.701 & 0.594 & 0.542  & 0.871 & 0.817 & 0.858 & 0.906 \\ 

\textbf{Mip-NeRF360} & 0.685 & 0.584 & 0.809 & 0.745 & 0.631 & 0.910 & 0.892 & 0.917 & 0.938\\

\hline
\textbf{Scaffold-GS$^*$} &  0.755&  0.590 & 0.858&0.765& 0.639 &0.923&0.910&0.926 & 0.943 \\
\textbf{Octree-GS$^*$} & 0.759& 0.597 &0.864&0.763&0.643&0.924&0.907&0.916&0.930 \\

\hline
\textbf{3D-GS} &0.771 & 0.605 & 0.868 & 0.775 & 0.638 & 0.914 & 0.905 & 0.922 & 0.938
 \\ 
\textbf{AbsGS} & 0.783 & 0.623 & 0.871 & 0.780 & 0.617 & 0.925 & 0.911 & \cellcolor{tabthird}0.929 & 0.945 \\ 

\textbf{Pixel-GS$^*$} & 0.775& 0.633& 0.867& 0.784& 0.630& 0.920& 0.911& \cellcolor{tabthird}0.929& 0.944 \\
\textbf{Mini-Splatting-D} & \cellcolor{tabfirst}0.798 & \cellcolor{tabsecond}0.642 & \cellcolor{tabfirst}0.878 & \cellcolor{tabfirst}0.804 & 0.640 & \cellcolor{tabthird}0.928 & 0.913 & \cellcolor{tabfirst}0.934 & \cellcolor{tabthird}0.946
 \\
\hline
\textbf{Ours-Small (AbsGS)} & \cellcolor{tabsecond}0.792& 0.634& 0.872& \cellcolor{tabsecond}0.802& \cellcolor{tabfirst}0.652& \cellcolor{tabsecond}0.929& \cellcolor{tabthird}0.914& \cellcolor{tabsecond}0.932& 0.945 \\
\textbf{Ours (3D-GS)} & \cellcolor{tabthird}0.790& \cellcolor{tabthird}0.639& \cellcolor{tabthird}0.874& \cellcolor{tabthird}0.800& \cellcolor{tabsecond}0.650& \cellcolor{tabsecond}0.929& \cellcolor{tabsecond}0.916& \cellcolor{tabfirst}0.934& \cellcolor{tabsecond}0.947 \\
\textbf{Ours (AbsGS)} & \cellcolor{tabfirst}0.797& \cellcolor{tabfirst}0.647& \cellcolor{tabsecond}0.876&  \cellcolor{tabfirst}0.804& \cellcolor{tabthird}0.645& \cellcolor{tabfirst}0.931& \cellcolor{tabfirst}0.918& \cellcolor{tabfirst}0.934& \cellcolor{tabfirst}0.948 \\
\bottomrule
\end{tabular}}
\end{table*}

\begin{table*}[]
\centering
\caption{\textbf{Per-scene LPIPS metrics of Mip-NeRF360 dataset.} 
}
\label{tab:mip_lpips}
\resizebox{0.86\linewidth}{!}{
\begin{tabular}{c|ccccc|cccc}
\toprule
\begin{tabular}{c|c} Method & Scenes \end{tabular}  & bicycle & flowers & garden & stump & treehill  & room & counter & kitchen & bonsai\\
\midrule
\textbf{Plenoxels} & 0.506 & 0.521 & 0.3864 & 0.503 & 0.540 & 0.4186 & 0.441 & 0.447 & 0.398\\ 

\textbf{Instant-NGP} & 0.446 & 0.441 & 0.257 & 0.421 & 0.450  & 0.261 & 0.306 & 0.195 & 0.205 \\ 

\textbf{Mip-NeRF360} & 0.301 & 0.344 & 0.170 & 0.261 & 0.339 & 0.211 & 0.204 & 0.127 & 0.176
\\

\hline
\textbf{Scaffold-GS$^*$} &  0.233&  0.352 &  0.120& 0.236& 0.331 & 0.216& 0.203& 0.130 &  0.207 \\
\textbf{Octree-GS$^*$} & 0.225& 0.340 & 0.108& 0.233& 0.310& 0.203& 0.202& 0.141& 0.219\\

\hline
\textbf{3D-GS} &0.205 & 0.336 & 0.103 & 0.210 & 0.317 & 0.220 & 0.204 & 0.129 & 0.205
 \\ 
\textbf{AbsGS} & 0.171 & 0.270 & 0.100 & 0.195 & \cellcolor{tabthird}0.278 & 0.200 & 0.189 & 0.121 & 0.190\\ 

\textbf{Pixel-GS$^*$} & 0.182& \cellcolor{tabthird}0.263& 0.100& 0.186& 0.280& 0.213& 0.185& 0.120& 0.193 \\
\textbf{Mini-Splatting-D} & \cellcolor{tabsecond}0.158 & \cellcolor{tabsecond}0.255 & \cellcolor{tabsecond}0.090 & \cellcolor{tabfirst}0.169 & \cellcolor{tabsecond}0.262 & \cellcolor{tabsecond}0.190 & \cellcolor{tabfirst}0.172 & \cellcolor{tabsecond}0.114 & \cellcolor{tabfirst}0.175 \\
\hline
\textbf{Ours-Small (AbsGS)} & 0.175& 0.290& 0.099& 0.188& 0.279& 0.196& \cellcolor{tabthird}0.184& \cellcolor{tabthird}0.117& 0.189\\
\textbf{Ours (3D-GS)} & \cellcolor{tabthird}0.168& 0.295& \cellcolor{tabthird}0.092& \cellcolor{tabthird}0.179& 0.280& \cellcolor{tabthird}0.192& \cellcolor{tabsecond}0.177& \cellcolor{tabsecond}0.114& \cellcolor{tabthird}0.185 \\
\textbf{Ours (AbsGS)} & \cellcolor{tabfirst}0.156& \cellcolor{tabfirst}0.251& \cellcolor{tabfirst}0.089& \cellcolor{tabsecond}0.170& \cellcolor{tabfirst}0.250& \cellcolor{tabfirst}0.187& \cellcolor{tabfirst}0.172& \cellcolor{tabfirst}0.112& \cellcolor{tabsecond}0.179 \\
\bottomrule
\end{tabular}}
\end{table*}